\documentclass{article}
\usepackage{ijcai24}

\usepackage{times}
\usepackage{soul}
\usepackage{url}
\usepackage[hidelinks]{hyperref}
\usepackage[utf8]{inputenc}
\usepackage[small]{caption}
\usepackage{graphicx}
\usepackage{amsmath}
\usepackage{amsthm}
\usepackage{booktabs}
\usepackage{algorithm}
\usepackage{algorithmic}
\usepackage[switch]{lineno}

\usepackage{amsfonts} 
\usepackage{subfloat}
\usepackage{subfig}

\title{Bridging the Gap: Learning Pace Synchronization for Open-World Semi-Supervised Learning}

\def\algo{{\textsc{LPS}}}

\author{
Bo Ye$^{1,2}$,
Kai Gan$^{1,2}$,
Tong Wei$^{1,2}$\footnotemark[2],
Min-Ling Zhang$^{1,2}$
\affiliations
$^1$School of Computer Science and Engineering, Southeast University, Nanjing 211189, China\\
$^2$Key Lab. of Computer Network and Information Integration (Southeast University), MoE, China
\emails
\{yeb, gank, weit, zhangml\}@seu.edu.cn
}

\begin{document}

\maketitle

\renewcommand{\thefootnote}{\fnsymbol{footnote}}
\footnotetext[2]{Corresponding author}

\begin{abstract}
    In open-world semi-supervised learning, a machine learning model is tasked with uncovering novel categories from unlabeled data while maintaining performance on seen categories from labeled data. The central challenge is the substantial learning gap between seen and novel categories, as the model learns the former faster due to accurate supervisory information. Moreover, capturing the semantics of unlabeled novel category samples is also challenging due to the missing label information. To address the above issues, we introduce 1) the adaptive synchronizing marginal loss which imposes class-specific negative margins to alleviate the model bias towards seen classes, and 2) the pseudo-label contrastive clustering which exploits pseudo-labels predicted by the model to group unlabeled data from the same category together in the output space. Extensive experiments on benchmark datasets demonstrate that previous approaches may significantly hinder novel class learning, whereas our method strikingly balances the learning pace between seen and novel classes, achieving a remarkable 3\% average accuracy increase on the ImageNet dataset. Importantly, we find that fine-tuning the self-supervised pre-trained model significantly boosts the performance, which is overlooked in prior literature. Our code is available at https://github.com/yebo0216best/LPS-main.
\end{abstract}

\section{Introduction}
Over the past decade, Semi-Supervised Learning (SSL) algorithms \cite{zhu2009introduction} have demonstrated remarkable performance across multiple tasks, even when presented with a meagre number of labeled training samples. These algorithms delve into the underlying data distribution by harnessing numerous unlabeled samples. Among the representative methods employed for this purpose are pseudo-labeling \cite{lee2013pseudo} and consistency regularization \cite{laine2016temporal,sajjadi2016regularization}. Pseudo-labeling involves utilizing model predictions as target labels, while consistency regularization encourages similar predictions for distinct views of an unlabeled sample. However, the majority of current approaches operate under the assumption that unlabeled data exclusively comprises samples belonging to \textit{seen classes}, as observed within the labeled data \cite{bendale2015towards,boult2019learning}. In contrast, the presence of samples from \textit{novel classes} in the unlabeled data is common, as it is challenging for human annotators to discern such instances amidst an extensive pool of unlabeled samples \cite{oliver2018realistic}. 

\begin{figure}[t]
\centering 
\includegraphics[width=\linewidth]{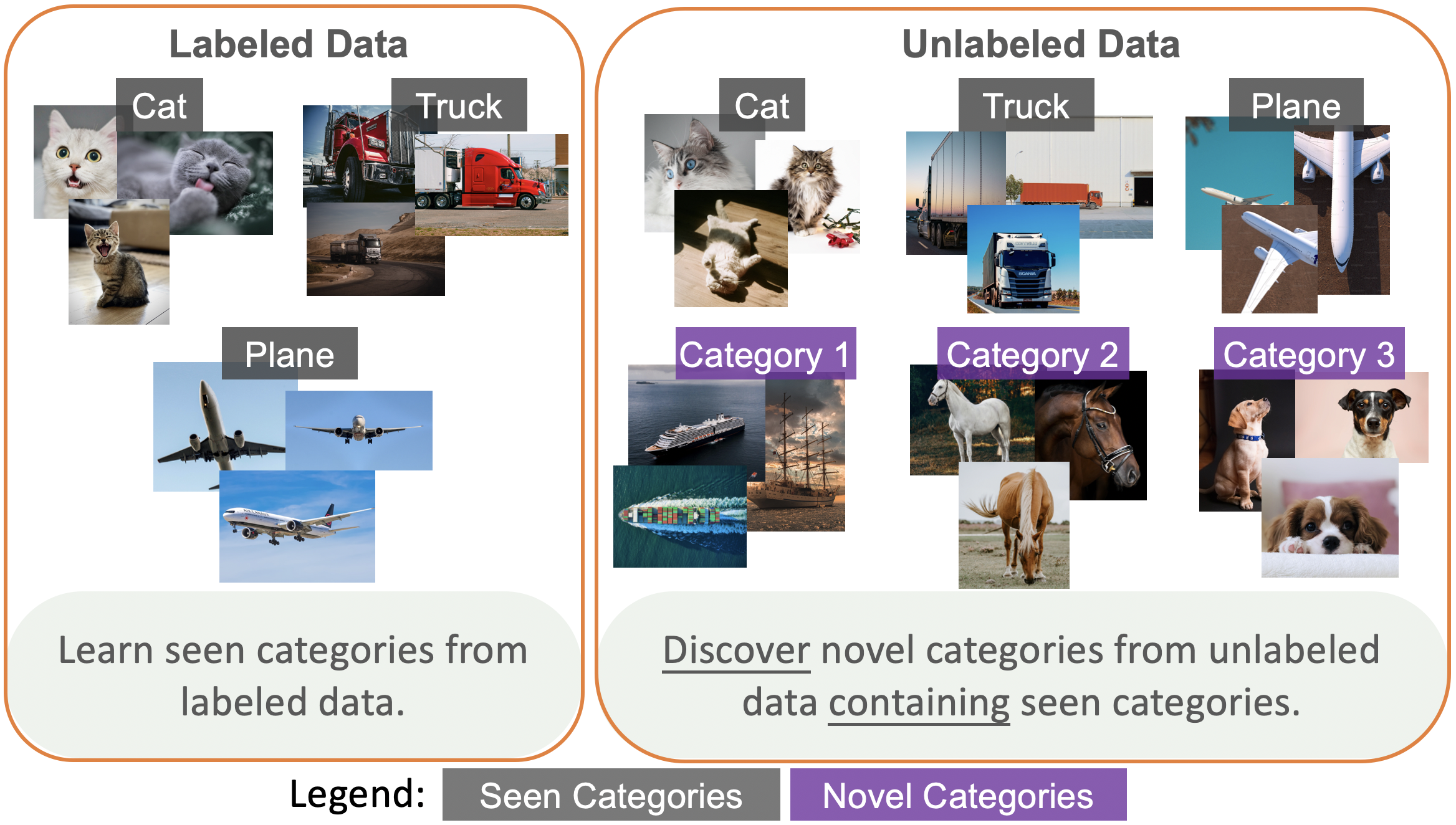}
\caption{Summary of OpenSSL setting.}\label{fig:setting}
\end{figure}

To aid this challenge, Open-World Semi-Supervised Learning, denoted as \textit{OpenSSL}, has gained recent attention, leading to the proposition of several effective methodologies \cite{cao2022openworld,guo2022nach,liu2023openncd}. Figure~\ref{fig:setting} demonstrates the problem setting of OpenSSL as an intuitive example. To tackle this issue, extant methods adopt a two-pronged strategy. On one front, they endeavour to identify unlabeled samples pertaining to seen classes and allocate pseudo-labels accordingly. On the other front, they automatically cluster unlabeled samples belonging to novel categories. Notably, OpenSSL shares an affinity with Novel Class Discovery (NCD) \cite{han2019learning,han2020automatically}, particularly concerning the clustering of novel class samples. However, NCD methodologies presuppose that unlabeled samples originate exclusively from novel classes. OpenSSL relaxes this assumption to mirror real-world scenarios 
more accurately. Evidently, the central challenge of effectively clustering novel class samples hinges upon the acquisition of discriminative feature representations, given the absence of supervisory information. To mitigate this quandary, existing methods harness self-supervised learning paradigms (e.g., SimCLR \cite{chen2020simple}) which circumvent the need for labeled data during the training of feature extractors within deep neural networks. Subsequently, a linear classifier is cultivated by optimizing the cross-entropy loss for labeled data, in conjunction with specifically tailored unsupervised objectives for the unlabeled counterpart. Widely employed unsupervised objectives include entropy regularization and pair-wise loss, both of which effectively enhance performance.

\begin{figure}[tp]
\centering 
\begin{minipage}{0.23\textwidth}
        \includegraphics[width=\linewidth]{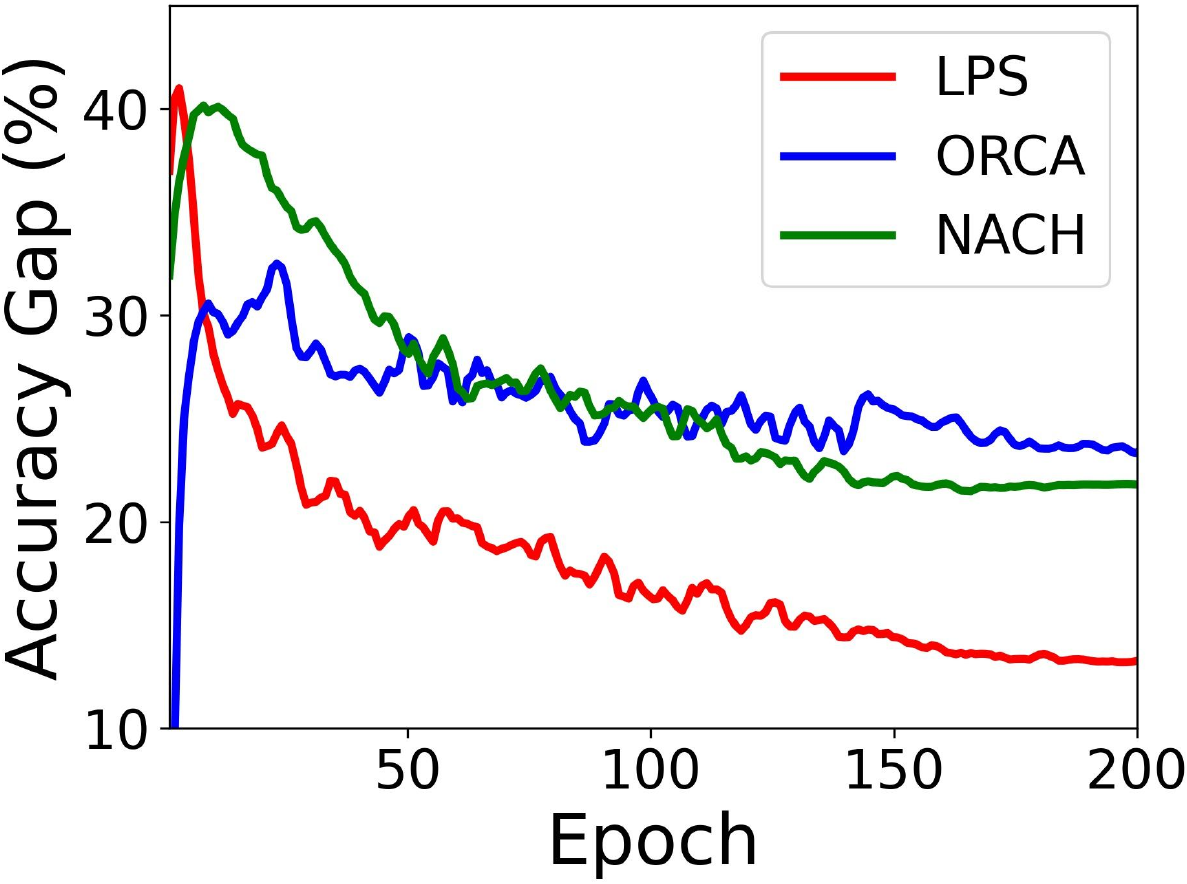}
        \caption*{(a) Accuracy Gap.}
\end{minipage}
\begin{minipage}{0.238\textwidth}
        \includegraphics[width=\linewidth]{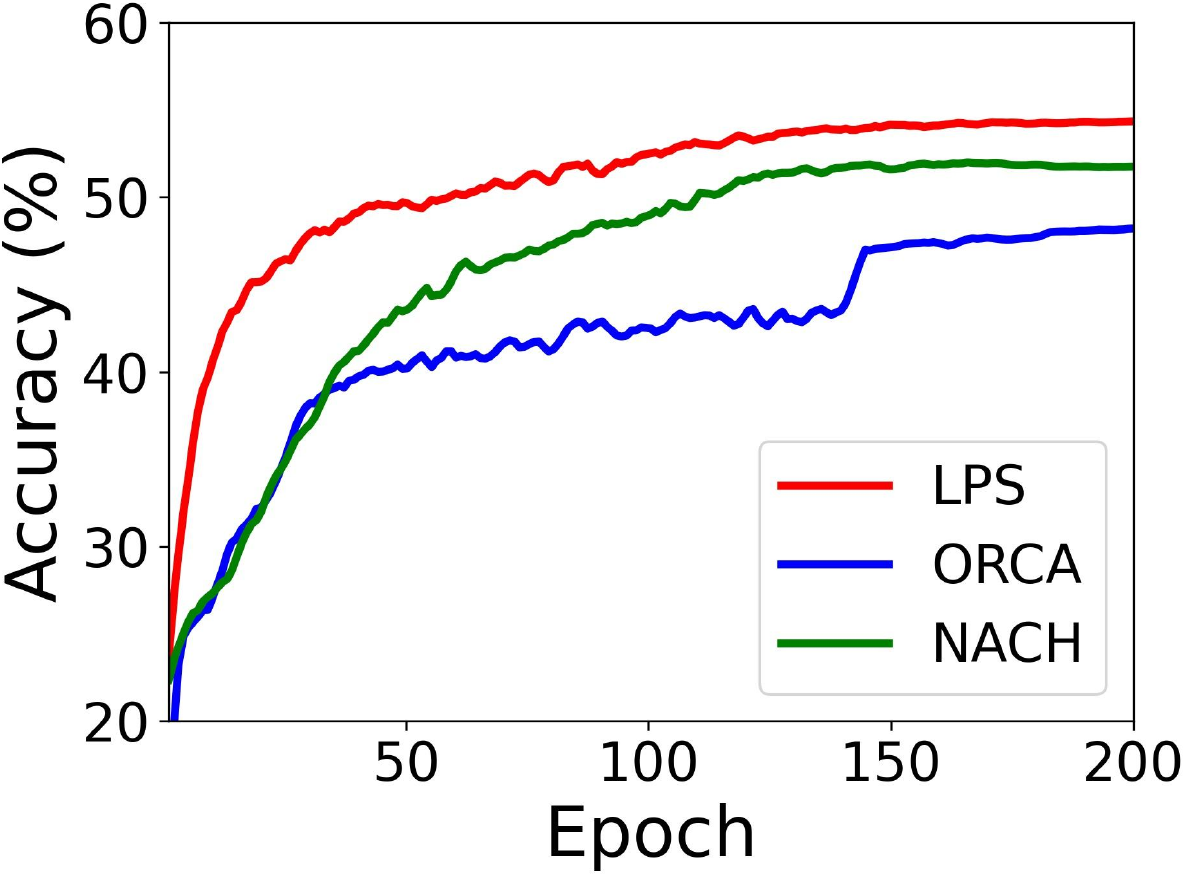}
        \caption*{(b) Overall Accuracy.}
\end{minipage}
\caption{(a) Accuracy gap between seen and novel classes. (b) The overall accuracy of LPS, ORCA, and NACH. Experiments are conducted on the CIFAR-100 dataset with 50\% seen classes (50\% labeled) and 50\% novel classes.}\label{fig:accgap}
\end{figure}

This paper introduces a novel OpenSSL algorithm. An initial observation reveals that the model exhibits faster learning of seen classes compared to novel classes. This discrepancy is intuitive because of accurate supervision within labeled data for seen classes, whereas novel classes are learned through unsupervised means. Figure \ref{fig:accgap} depicts a demonstration of the learning speed discrepancy between seen and novel classes. Motivated by this intrinsic problem, we propose an adaptive distribution-aware margin mechanism, designed to steer the model's attention towards novel class learning. Notably, this margin diminishes as the model's predicted class distribution approaches the underlying (class-balanced) distribution. To learn robust representations and facilitate the clustering of novel classes, we introduce pseudo-label contrastive clustering. This technique aggregates unlabeled samples sharing the same class, guided by model predictions. Importantly, we exploit multiple positive and negative pairs as supervisory signals, in contrast to the reliance on a single positive pair as seen in previous works. For unlabeled samples exhibiting low confidence, we integrate unsupervised contrastive learning to facilitate the acquisition of informative representations. This unsupervised contrastive objective operates as a complement to the pseudo-label contrastive clustering. Combining the aforementioned modules, we present, \algo, to address the OpenSSL challenge. Figure~\ref{fig:accgap} showcases the efficacy of \algo\ compared with existing state-of-the-art approaches. Notably, we reveal that the conventional practice of freezing the feature extractor, previously trained via self-supervised learning in prior research, falls short of optimal.

In summary, our main contributions are:
\begin{itemize}
\item  We propose a novel and simple method, \algo, to effectively synchronize the learning pace of seen and novel classes for open-world semi-supervised learning.
\item We conduct extensive experiments to verify the effectiveness of the proposed method against the previous state-of-the-art. Particularly, \algo\ achieves over 3\% average increase of accuracy on the ImageNet dataset.
\item We examine the effectiveness of the key components of the proposed method. Different from previous works, we discover that fine-tuning the pre-trained backbone allows the model to learn more useful features, which can significantly improve the performance.
\end{itemize}

\section{Related Work}
\smallskip
\noindent\textbf{Semi-Supervised Learning.} 
Within the realm of SSL, pseudo-labeling \cite{lee2013pseudo} and consistency regularization \cite{laine2016temporal,sajjadi2016regularization,DBLP:journals/corr/abs-2205-13358,DBLP:conf/cvpr/WeiG23} are two widely used techniques. Pseudo-labeling converts model predictions on unlabeled samples into either soft labels or hard labels, subsequently employed as target labels. Consistency regularization strives to ensure model outputs exhibit a high degree of consistency when applied to perturbed samples. Recent advancements \cite{berthelot2019mixmatch,sohn2020fixmatch,xu2021dash} combine pseudo-labeling with consistency regularization to yield further performance enhancements. In addition to the above techniques, the application of contrastive learning into SSL has also received substantial interest. For example, TCL \cite{singh2021tcll} introduces contrastive learning as a tool to enhance representation learning. TCL maximizes agreement between different views of the same sample while minimizing agreement for distinct samples. In consonance with this paradigm, we design a new complementary contrastive loss to explore the consistency of all samples effectively.

\smallskip
\noindent\textbf{Novel Class Discovery.} 
The setting of NCD aligns closely with the scenario investigated in this paper. NCD  assumes a scenario where the labeled data consists of samples of seen classes, while the unlabeled data exclusively comprises samples of novel classes. \cite{han2019learning} initially raised the NCD problem. Subsequent research such as \cite{han2020automatically,zhong2021neighborhood,zhong2021openmix} predominantly adopted multi-stage training strategies. The underlying principle is capturing comprehensive high-level semantic information from labeled data, subsequently propagated to unlabeled counterparts. The majority of NCD methods involve preliminary model pre-training, wherein several objective functions are invoked to minimize inter-sample distances for each class. However, in real-world scenarios, the assumption of unlabeled data solely comprising novel classes is unrealistic, as seen classes also significantly populate the unlabeled dataset. Our experimentation reveals that NCD algorithms struggle to match the performance of other leading methods in the context of OpenSSL.

\begin{figure*}[t]
    \centering
    \includegraphics[width=0.9\textwidth]{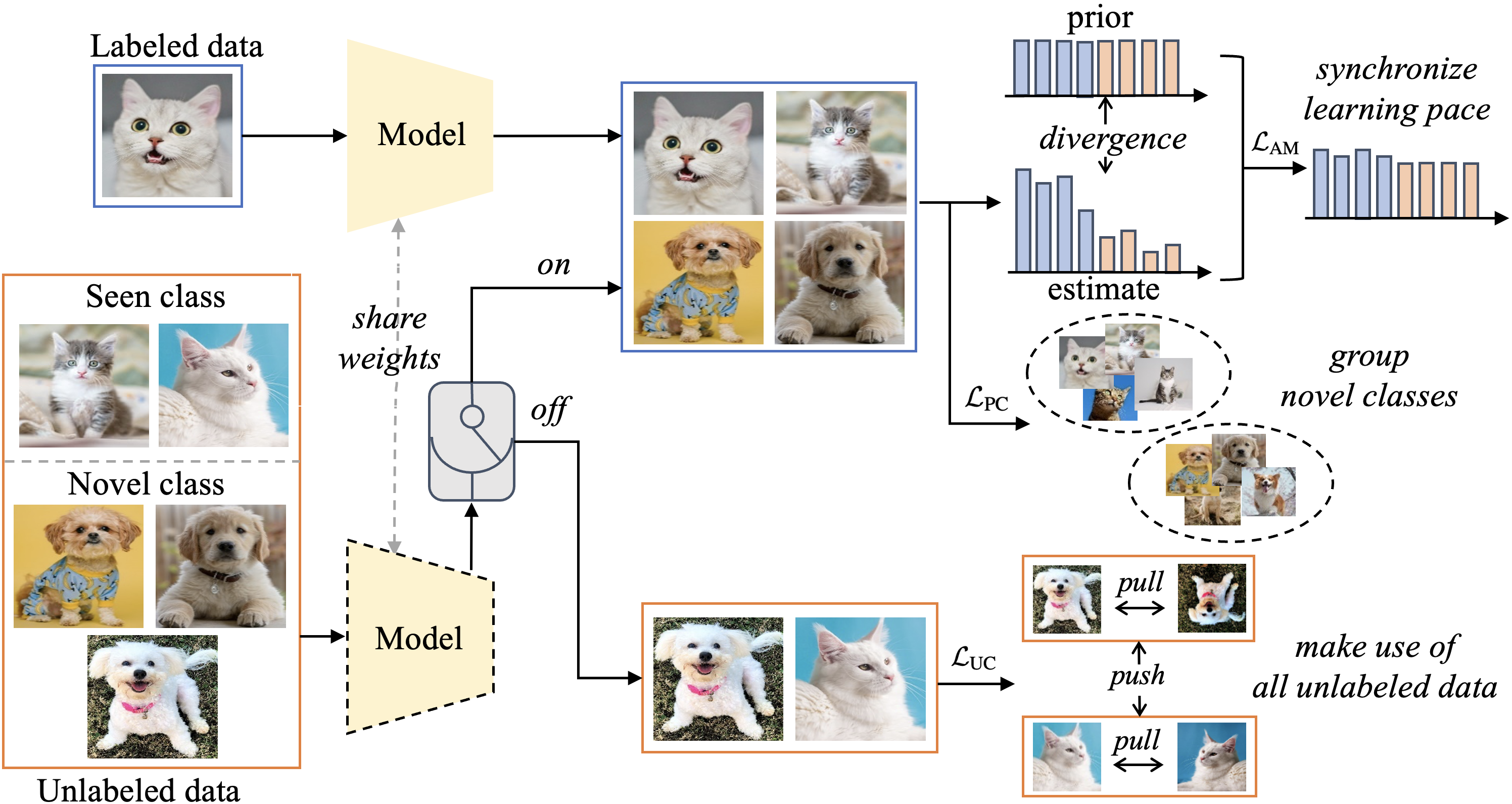}
    \caption{Overview of LPS framework. The LPS objective function is composed of an adaptive margin objective $\mathcal{L}_{\text{AM}}$, a pseudo-label contrastive clustering objective $\mathcal{L}_{\text{PC}}$, and an unsupervised contrastive learning objective $\mathcal{L}_{\text{UC}}$.}
    \label{fig:model}\label{fig:framework}
\end{figure*}

\smallskip
\noindent\textbf{Open-World Semi-Supervised Learning.} 
While conventional SSL methods operate under the assumption of labeled and unlabeled data being associated with a predefined set of classes, recent advancements \cite{oliver2018realistic,chen2020semi,guo2020safe,saito2021openmatch} challenge this notion and acknowledge the potential emergence of novel classes within the unlabeled data. Further, ORCA \cite{cao2022openworld} delves into the OpenSSL problem and attempts to cluster novel classes from unlabeled data. Both ORCA and NACH \cite{guo2022nach} notice the discrepancy in learning pace between seen and novel classes. They propose to use an entropy regularizer and pair-wise loss to ameliorate this issue. More recently, OpenNCD \cite{liu2023openncd} introduces prototypes-based contrastive learning for recognizing seen classes and discovering novel classes. This paper introduces an adaptive margin loss and pseudo-label contrastive clustering mechanism to synchronize the learning pace and enhance novel class discovery.

\section{The Proposed Method}
\paragraph{Notations.} The training dataset is composed of the labeled data $\mathcal{D}_l = \{(\boldsymbol{x_i},y_i)\}_{i=1}^n$ and the unlabeled data $\mathcal{D}_u = \{\boldsymbol{x_{i}}\}_{i=n+1}^{n+m}$. Within the context of OpenSSL, the classes in $\mathcal{D}_l$ are designated as \textit{seen classes}, constituting the set denoted as $\mathcal{C}_s$. The scenario of interest acknowledges a distribution mismatch, leading to $\mathcal{D}_u$ comprising instances from both seen and \textit{novel classes}. The collection of these novel classes is represented as $\mathcal{C}_n$. Additionally, we adopt the premise of a known number of novel classes akin to prior OpenSSL methodologies \cite{cao2022openworld,guo2022nach,liu2023openncd}. The goal of OpenSSL is to classify samples originating from the $\mathcal{C}_s$ and cluster samples emanating from the $\mathcal{C}_n$.

\paragraph{Overview.}
The fundamental challenge in OpenSSL arises from the pronounced discrepancy in learning paces between seen and novel classes, primarily due to the precise supervisory guidance for seen classes. This discrepancy results in a bias towards seen classes in the model's predictions, adversely impacting both the accurate classification of seen class samples and the effective clustering of novel class samples. To circumvent this challenge, we introduce \textit{Learning Pace Synchronization} (LPS), a methodology with adaptive synchronizing loss and pseudo-label contrastive clustering as in Figure \ref{fig:framework}. The adaptive synchronizing loss aims to achieve a balance between the learning pace of seen and novel classes, and the pseudo-label contrastive clustering exploits pseudo-labels to group unlabeled data from the same class together in the output space.


\subsection{Adaptive Synchronizing}
To start with, we describe the proposed adaptive marginal loss which regularizes the learning pace of seen classes to synchronize the learning pace of the model. Conventionally, the margin is defined as the minimum distance of the data to the classification boundary. For a sample $(\boldsymbol{x}, y)$, we have:
\begin{equation}
\Delta(\boldsymbol{x}, y)=f(\boldsymbol{x})_y-\max _{j \neq y} f(\boldsymbol{x})_j
\end{equation}
Instead of employing a fixed margin, LDAM \cite{cao2019learning} introduces a class-specific margin, where the margin between rare classes and other classes is larger than the margin for frequent classes, for tackling class-imbalanced data. Specifically, it sets the margin of class $j$ as:
\begin{equation}\label{equ:ldam}
\Delta_j=\frac{C}{n_j^{1 / 4}}
\end{equation}
The constant $C$ controls the intensity and $n_j$ denotes the frequency of class $j$ in the training data. Motivated by this, we propose a new variant of the margin loss to synchronize the learning pace of seen and novel classes. We apply adaptive margin loss to demand a larger margin between the novel and other classes, so that scores for seen classes, towards which the model highly biased, do not overwhelm the novel classes.
For each sample ($\boldsymbol{x}, y$), the adaptive margin loss is defined as follows:
\begin{equation}\label{equ:am}
\begin{aligned}
&\ell_{\text{AM}}(\boldsymbol{x},y)  = -\log \frac{\text{exp}({z_{y}-\Delta_{y}})}{\text{exp}({z_{y}-\Delta_{y}}) + \sum_{j \neq y} \text{exp}({z_j})} \\ 
&\textup{where }{\Delta_j = - KL\Big(\boldsymbol{\widehat{\pi}}\ \big\| \ \boldsymbol{\pi} \Big)\frac{{\widehat{\pi}_j}}{\max(\boldsymbol{\boldsymbol{\widehat{\pi}}})}C, \;\; j \in [K]} 
\end{aligned}
\end{equation}
In this formulation, $K$ represents the total number of classes, $z_{j}$ signifies the model output for the $j$-th class, $\boldsymbol{z}=f(\boldsymbol{x};\theta)$, and $\boldsymbol{\widehat{\pi}}$ denotes the estimated class distribution by the model. Additionally, we introduce an approximation of the true class distribution $\boldsymbol{\pi}$, which is naturally inaccessible during training. In line with prior studies, we assume a uniform distribution for $\boldsymbol{\pi}$, leaving the exploration of arbitrary distributions for future investigations. The hyper-parameter $C$ is introduced to control the maximum margin, and we empirically set $C=10$ across all experiments. We conduct a series of studies on the value of $C$ in the supplementary material. 

For the sake of simplicity, we assume that the mini-batch is comprised of labeled data ${B_l}$ and unlabeled data ${B_u}$. Given that the computation of Eq. \eqref{equ:am} relies on the class distribution, we proceed to estimate the complete class distribution through labeled data and unlabeled data exhibiting high predictive confidence. Specifically, we endeavour to achieve this estimation through:
\begin{equation}
\begin{aligned}
\!\!\boldsymbol{\widehat{\pi}}= \text{Normalize}\!\left(\!\sum_{\boldsymbol{x}_i \in {B}_l} \widehat{\boldsymbol{y}}_i\! +\!\!\!\!\sum_{\boldsymbol{x}_j \in {B}_u}\!\!\mathbb{I}\left(\max(\widehat{\boldsymbol{y}}_j)\! \geq \!\lambda\right) \widehat{\boldsymbol{y}}_j\!\!\right)
\end{aligned}
\end{equation}
Here, $\widehat{\boldsymbol{y}}=\text{softmax}(\boldsymbol{z})$. In view of the tendency for novel class samples to exhibit underconfidence, we empirically introduce a progressively evolving confidence threshold ${\lambda}_{novel}=0.4+0.4 \times \frac{t}{T}$, where ${t}$ and ${T}$ signify the current training iteration and the total training iterations, respectively. For seen classes, a fixed confidence threshold  ${\lambda}_{seen}=0.95$ is employed.

Given that the estimated class distribution $\boldsymbol{\widehat{\pi}}$ mirrors the model's confidence in class predictions, we harness this insight to regulate the learning pace of both seen and novel classes. Notably, in the early training phases, the model is inclined towards seen classes, with the logit adjustment term $\Delta_j$ assuming a larger negative margin for seen classes, thereby attenuating their learning pace. As training progresses, the model attains a more balanced capability across both seen and novel classes, as reflected by a diminishing Kullback-Leibler (KL) divergence between $\boldsymbol{\widehat{\pi}}$ and $\boldsymbol{\pi}$.

In summary, the adaptive margin loss $\mathcal{L}_{\text {AM}}$ for both labeled and pseudo-labeled data is defined as follows:
\begin{equation}
\begin{aligned}
\mathcal{L}_{\text {AM}}= & \frac{1}{|B_{l}|}\sum_{\boldsymbol{x}_i \in {B}_l}\ell_{\text{AM}}(\boldsymbol{z}^w_i,y) + \\
& \frac{1}{|B_{u}|}\sum_{\boldsymbol{x}_j \in {B}_u}\!\!\!\mathbb{I}\left(\max(\widehat{\boldsymbol{y}}_j) \geq \!\lambda\right)\ell_{\text{AM}}(\boldsymbol{z}^s_j,\widehat{p}_j)
\end{aligned}
\end{equation}
In this context, $\boldsymbol{z}^w$ and $\boldsymbol{z}^s$ correspond to the output logits stemming from the weak and strong augmented versions of sample $\boldsymbol{x}$, respectively. The symbol $|\cdot|$ denotes the set cardinality operation. Additionally, we utilize $\widehat{p} = \arg\max(\text{softmax}(\boldsymbol{z}^w))$ to represent the pseudo-label associated with the sample. 

\paragraph{Distinctions and Connections with Alternatives.} It is worth noting that the concept of adaptive margin has been used in prior literature \cite{li2020aml,mai2021DRaml,cao2022openworld}. Different from LPS, \cite{li2020aml} leverages the semantic similarities between classes to generate adaptive margins with the motivation to separate similar classes in the embedding space, and \cite{mai2021DRaml} utilizes the ground-truth distance between different samples to generate adaptive margins with the motivation to adapt to rating datasets. 
In OpenSSL, ORCA \cite{cao2022openworld} also integrates an adaptive margin mechanism based on the model's predictive uncertainty, which can only equally suppress the learning pace of seen classes. However, there are still differences in the learning paces of different classes among seen classes. Our proposed adaptive margin is based on the current estimated distribution to reflect the learning pace of different classes, which offers increased flexibility for regulating the learning pace across classes by generating the class-specific negative margin. Furthermore, the inclusion of the KL divergence term effectively guards against the model converging to a trivial solution where all samples are arbitrarily assigned to a single class.

\subsection{Pseudo-Label Contrastive Clustering}
The basic idea of discovering novel classes is to explore the correlations between different samples and cluster them into several groups. Prior OpenSSL approaches often transform the clustering task into a pairwise similarity prediction task, wherein a modified form of binary cross-entropy loss is optimized. Different from existing works, we introduce a new clustering method to fully exploit reliable model predictions as supervisory signals.


Our approach involves the construction of a multi-viewed mini-batch by using weak and strong augmentations. Within each mini-batch, we group the labeled and confident unlabeled samples, which is denoted as $B_{l'}$. Concurrently, unlabeled samples exhibiting predicted confidence levels failing below the threshold $\lambda$ are denoted as $B_{u'}$. Pseudo-label contrastive clustering only takes $B_{l'}$ as inputs. For each sample in $B_{l'}$, the set of the positive pairs contains samples with the same given label or pseudo-label. Conversely, the set of negative pairs contains samples of other classes. Formally, the objective of pseudo-label contrastive clustering is defined as follows:
\begin{equation}\label{equ:pc}
  \mathcal{L}_\text{PC}\!=\!-\frac{1}{|B_{l'}|}\!\!\sum_{\boldsymbol{x}_i\in B_{l'}}\!\log\!\frac{1}{|P(\boldsymbol{x}_i)|}\frac{\!\!\!\sum\limits_{\boldsymbol{x}_p\in P(\boldsymbol{x}_i)}\!\!\text{exp}\left(\boldsymbol{z}_i\!\cdot\!\boldsymbol{z}_p/\tau\right)}{\!\sum\limits_{\boldsymbol{x}_a\in A(\boldsymbol{x}_i)}\!\!\!\!\text{exp}\left(\boldsymbol{z}_i\!\cdot\!\boldsymbol{z}_a/\tau\right)},
\end{equation}
where $\boldsymbol{z}_i$ denotes the output logits, $P(\boldsymbol{x}_i)$ denotes the set of positives of $\boldsymbol{x}_i$ and $A(\boldsymbol{x}_i)\equiv B_{l'}\backslash \{\boldsymbol{x}_i\}$. $\tau$ is a tunable temperature parameter and we set $\tau=0.4$ in all experiments.

In contrast to existing methods such as ORCA \cite{cao2022openworld} and NACH \cite{guo2022nach}, which establish a single positive pair for each sample by identifying its nearest neighbour, the objective in Eq. \eqref{equ:pc} adopts pseudo-labels to form multiple positive pairs. This approach offers dual advantages: firstly, the alignment of samples within the same class is more effectively harnessed through the utilization of multiple positive sample pairs; secondly, it leverages the consistency of distinct views of the same sample to mitigate the negative impact of erroneous positive pairs, while concurrently imparting a repulsion effect to samples from different classes through negative pairs.

\begin{table*}[ht]
\resizebox{\textwidth}{!}{%
\setlength{\tabcolsep}{1.3mm}{
\begin{tabular}{@{}lcccccccccccccccccc@{}}
\toprule
  \multicolumn{1}{l}{} &
  \multicolumn{6}{c}{\textbf{CIFAR-10}} &
  \multicolumn{6}{c}{\textbf{CIFAR-100}} &
  \multicolumn{6}{c}{\textbf{ImageNet-100}} \\ \midrule
\multicolumn{1}{c}{} &
  \multicolumn{3}{c}{\textbf{10\% labeled}} &
  \multicolumn{3}{c}{\textbf{50\% labeled}} &
  \multicolumn{3}{c}{\textbf{10\% labeled}} &
  \multicolumn{3}{c}{\textbf{50\% labeled}} &
  \multicolumn{3}{c}{\textbf{10\% labeled}} &
  \multicolumn{3}{c}{\textbf{50\% labeled}} \\ \cmidrule(lr){2-4} \cmidrule(lr){5-7} \cmidrule(lr){8-10} \cmidrule(lr){11-13} \cmidrule(lr){14-16} \cmidrule(l){17-19} 
\textbf{Methods} &
  \textbf{Seen} &
  \textbf{Novel} &
   \textbf{All} &
  \textbf{Seen} &
  \textbf{Novel} &
   \textbf{All} &
  \textbf{Seen} &
  \textbf{Novel} &
   \textbf{All} &
  \textbf{Seen} &
  \textbf{Novel} &
   \textbf{All} &
  \textbf{Seen} &
  \textbf{Novel} &
   \textbf{All} &
  \textbf{Seen} &
  \textbf{Novel} &
   \textbf{All} \\ \midrule
\begin{tabular}[c]{@{}l@{}}FixMatch\\ DS$^3$L\\ DTC\\ RankStats\\ SimCLR\\ ORCA\\ GCD\\ OpenNCD\\ NACH\\  \algo\ (ours)\end{tabular} &
  \begin{tabular}[c]{@{}c@{}}64.3\\ 70.5\\ 42.7\\ 71.4\\ 44.9\\ 82.8\\ 78.4\\ 83.5\\ 86.4\\ 86.3\end{tabular} &
  \begin{tabular}[c]{@{}c@{}}49.4\\ 46.6\\ 31.8\\ 63.9\\ 48.0\\ 85.5\\ 79.7\\ 86.7\\ 89.4\\ 90.6\end{tabular} &
  \begin{tabular}[c]{@{}c@{}} 47.3\\  43.5\\  32.4\\  66.7\\  47.7\\  84.1\\  79.1\\  85.3\\  88.1\\  \textbf{88.6}\end{tabular} &
  \begin{tabular}[c]{@{}c@{}}71.5\\ 77.6\\ 53.9\\ 86.6\\ 58.3\\ 88.2\\ 78.4\\ 88.4\\ 89.5\\ 90.2\end{tabular} &
  \begin{tabular}[c]{@{}c@{}}50.4\\ 45.3\\ 39.5\\ 81.0\\ 63.4\\ 90.4\\ 79.7\\ 90.6\\ 92.2\\ 93.4\end{tabular} &
  \begin{tabular}[c]{@{}c@{}} 49.5\\  40.2\\  38.3\\  82.9\\  51.7\\  89.7\\  79.1\\  90.1\\  91.3\\  \textbf{92.4}\end{tabular} &
  \begin{tabular}[c]{@{}c@{}}30.9\\ 33.7\\ 22.1\\ 20.4\\ 26.0\\ 52.5\\ 49.7\\ 53.6\\ 57.4\\ 55.2\end{tabular} &
  \begin{tabular}[c]{@{}c@{}}18.5\\ 15.8\\ 10.5\\ 16.7\\ 28.8\\ 31.8\\ 27.6\\ 33.0\\ 37.5\\ 41.0\end{tabular} &
  \begin{tabular}[c]{@{}c@{}} 15.3\\  15.1\\  13.7\\  17.8\\  26.5\\  38.6\\  38.0\\  41.2\\  43.5\\  \textbf{47.5}\end{tabular} &
  \begin{tabular}[c]{@{}c@{}}39.6\\ 55.1\\ 31.3\\ 36.4\\ 28.6\\ 66.9\\ 68.5\\ 69.7\\ 68.7\\ 64.5\end{tabular} &
  \begin{tabular}[c]{@{}c@{}}23.5\\ 23.7\\ 22.9\\ 28.4\\ 21.1\\ 43.0\\ 33.5\\ 43.4\\ 47.0\\ 49.9\end{tabular} &
  \begin{tabular}[c]{@{}c@{}} 20.3\\  24.0\\  18.3\\  23.1\\  22.3\\  48.1\\  45.2\\  49.3\\  52.1\\  \textbf{54.3}\end{tabular} &
  \begin{tabular}[c]{@{}c@{}}60.9\\ 64.3\\ 24.5\\ 41.2\\ 42.9\\ 83.9\\ 82.3\\ 84.0\\ 86.3\\ 87.0\end{tabular} &
  \begin{tabular}[c]{@{}c@{}}33.7\\ 28.1\\ 17.8\\ 26.8\\ 41.6\\ 60.5\\ 58.3\\ 65.8\\ 66.5\\ 73.6\end{tabular} &
  \begin{tabular}[c]{@{}c@{}} 30.2\\  25.9\\  19.3\\  37.4\\  41.5\\  69.7\\  68.2\\  73.2\\  71.0\\  \textbf{78.0}\end{tabular} &
  \begin{tabular}[c]{@{}c@{}}65.8\\ 71.2\\ 25.6\\ 47.3\\ 39.5\\ 89.1\\ 82.3\\ 90.0\\ 91.0\\ 91.3\end{tabular} &
  \begin{tabular}[c]{@{}c@{}}36.7\\ 32.5\\ 20.8\\ 28.7\\ 35.7\\ 72.1\\ 58.3\\ 77.5\\ 75.5\\ 81.3\end{tabular} &
  \begin{tabular}[c]{@{}c@{}} 34.9\\  30.8\\  21.3\\  40.3\\  36.9\\  77.8\\  68.2\\  81.6\\  79.6\\  \textbf{84.5}\end{tabular} \\ \bottomrule
\end{tabular}%
}
}
    
    \caption{Accuracy comparison of seen, novel, and all classes on CIFAR-10, CIFAR-100, and ImageNet-100 datasets with 50\% classes as seen and 50\% classes as novel. We conducted experiments with 10\% and 50\% labeled data of seen classes.}
    \label{tab:10per_50per_label}
\end{table*}

Since Eq. \eqref{equ:pc} augments the labeled dataset by unlabeled samples of high predictive confidence, we ask whether unlabeled samples of low confidence can be used to enrich representation learning. In pursuit of this, we incorporate unsupervised contrastive learning \cite{wang2020understanding} to encourage similar predictions rather than embeddings between a given sample and its augmented counterpart. This helps to signify the uniformity among unlabeled samples, ultimately leading to clearer separations. In detail, for each sample in the low-confidence set $B_{u'}$, the unsupervised contrastive learning couples it with its augmented view to constitute a positive pair. Simultaneously, a set of negative pairs is formulated, containing all the samples within the mini-batch except the sample itself. The unsupervised contrastive learning loss $\mathcal{L}_{\text{UC}}$ is formulated as follows:
\begin{equation}\label{equ:uc}
\mathcal{L}_{\text{UC}}=-\frac{1}{|B_{u'}|}\sum_{\boldsymbol{x}_j\in B_{u'}}\log{\frac{\text{exp}\left(\boldsymbol{z}_j\cdot\boldsymbol{z}_{p}/\tau\right)}{\sum\limits_{\boldsymbol{x}_a\in \ A(\boldsymbol{x}_j)}\text{exp}\left(\boldsymbol{z}_j\cdot\boldsymbol{z}_a/\tau\right)}}
\end{equation}
Here, $\boldsymbol{x}_{p}$ is the positive sample of $\boldsymbol{x}_{j}$ and $A(\boldsymbol{x}_j)\equiv B_{u'}\cup B_{l'}\backslash \{\boldsymbol{x}_j\}$ for the sample $\boldsymbol{x}_j$.

In essence, unsupervised contrastive learning complements the pseudo-label contrastive clustering by fully exploiting the unlabeled samples. In the experiments, the ablation studies underscore the pivotal role played by both types of contrastive losses in our approach.

Lastly, we incorporate a maximum entropy regularizer to address the challenge of converging during the initial training phases, when the predictions are mostly wrong (e.g., the model tends to assign all samples to the same class) \cite{Arazo2020ijcnn}. Specifically, we leverage the KL divergence between the class distribution predicted by the model and a uniform prior distribution. It is worth noting that the integration of an entropy regularizer is a widespread practice in dealing with the OpenSSL problem, including approaches such as ORCA, NACH, and OpenNCD.
The final objective function of \algo\ is articulated as follows:
\begin{equation}
\mathcal{L}_{\text{total}} = \mathcal{L}_{\text{AM}}+\eta_1\mathcal{L}_{\text{PC}}+\eta_2\mathcal{L}_{\text{UC}}+\mathcal{R}_{\text{Entropy}}
\end{equation}
where $\mathcal{R}_{\text{Entropy}}$ denotes the entropy regularizer, $\eta_1$ and $\eta_2$ are hyper-parameters set to 1 in all our experiments. We provide detailed analyses on the sensitivity of hyperparameters in the supplementary material. 

\section{Experiments}
\subsection{Experimental Setup}
\smallskip
\noindent\textbf{Datasets.}
We evaluate our method on three commonly used datasets, i.e., CIFAR-10, CIFAR-100 \cite{krizhevsky2009learning}, and ImageNet \cite{russakovsky2015imagenet}. Following prior works \cite{cao2022openworld,guo2022nach,liu2023openncd}, we assume that the number of novel classes is known. Specifically, we randomly select 50\% of the classes as seen classes, and the remaining classes are regarded as novel classes, e.g., the number of novel classes is 50 for CIFAR-100. On each dataset, we consider two types of labeled ratios, i.e., only 10\% or 50\% of data in seen classes are labeled. For the ImageNet dataset, we subsample 100 classes to form the ImageNet-100 dataset for fair comparisons with existing works.

\smallskip
Following prior works \cite{cao2022openworld,guo2022nach,liu2023openncd}, we evaluate our method with respect to the accuracy of seen classes, novel classes, and all classes. For seen classes, the accuracy is calculated as the normal classification task. For novel classes, we first utilize the Hungarian algorithm \cite{kuhn55} to solve the optimal prediction-target class assignment problem and then calculate the accuracy of novel classes. For overall accuracy, we also solve the optimal assignment in the entire unlabeled dataset to calculate the novel class accuracy, measuring the overall performance. 

\smallskip
\noindent\textbf{Implementation Details.}
Following \cite{cao2022openworld,guo2022nach,liu2023openncd}, we utilize the self-supervised learning method SimCLR \cite{chen2020simple} to pre-train the backbone and fix the first three blocks. In LPS, the weak augmentation contains random crop and horizontal flip, and the strong augmentation is RandAugment \cite{cubuk2019randaugment}. For CIFAR-10 and CIFAR-100, we utilize ResNet-18 as our backbone which is trained by the standard SGD with a momentum of 0.9 and a weight decay of 0.0005. We train the model for 200 epochs with a batch size of 512. For the ImageNet dataset, we opt for ResNet-50 as our backbone. This choice also undergoes training via the standard SGD, featuring a momentum coefficient of 0.9 and a weight decay of 0.0001. The training process spans 90 epochs, with a batch size of 512. and The cosine annealing learning rate schedule is adopted on CIFAR and ImageNet datasets. These experiments are conducted on a single NVIDIA 3090 GPU. 

\subsection{Comparing with Existing Methods}
\smallskip
\noindent\textbf{Baselines.}
We compare LPS with SSL methods, open-set SSL methods, NCD methods, and OpenSSL methods. 
The NCD methods consider that the labeled data only has disjoint classes compared with the unlabeled data and aim at clustering novel classes without recognizing seen classes. For novel classes, clustering accuracy can be obtained directly. For seen classes, we first regard them as novel classes and leverage the Hungarian algorithm to match some of the discovered classes with seen classes, and then calculate the classification accuracy. We select two competitive NCD methods DTC \cite{han2019learning} and RankStats \cite{han2020automatically} in the experiments. Moreover, we include GCD \cite{vaze2022gcd} for comparison, which is an extended NCD method.

For the SSL and open-set SSL methods, we leverage their capability in estimating out-of-distribution samples to extend to the OpenSSL setting. For comparison, we select FixMatch\cite{sohn2020fixmatch}, which assigns pseudo-labels to unlabeled samples based on confidence. The classification accuracy of seen classes can be reported directly according to pseudo-labels. For novel classes, we first estimate samples without pseudo-labels as novel classes and then utilize $k$-means to cluster them. The open-set SSL methods maintain the classification performance of seen classes by rejecting novel classes. We compare with $\text{DS}^3\text{L}$ \cite{guo2020safe} and calculate its accuracy in the same way as FixMatch.

For the OpenSSL methods, we compare with ORCA \cite{cao2022openworld}, NACH \cite{guo2022nach}, and OpenNCD \cite{liu2023openncd}. We also compare the self-supervised pre-trained model SimCLR and conduct K-means on the primary features to calculate the accuracy.

\smallskip
\noindent\textbf{Results.}
The results on three datasets are reported in Table \ref{tab:10per_50per_label}. The mean accuracy is computed over three runs for each method. Although the non-OpenSSL methods perform well on their original tasks, their overall performance is unsatisfactory in the OpenSSL setting. The results of SimCLR are obtained by the pre-trained model without extra fine-tuning, and the OpenSSL methods are based on the pre-trained model. It is obvious that the OpenSSL methods achieve significant performance improvements compared to non-OpenSSL methods. Compared with the state-of-the-art OpenSSL methods, our method LPS achieves the best overall performance across all datasets. On the CIFAR-10 dataset, LPS outperforms NACH by 1.2\% in novel class accuracy. Likewise, on the CIFAR-100 dataset, LPS demonstrates superiority, yielding a substantial 3.2\% improvement. Particularly concerning the ImageNet-100 dataset, LPS has the capacity to surpass existing state-of-the-art methods, resulting in a 3.8\% increase in overall accuracy. Experimental results demonstrate that LPS can effectively balance the learning of seen and novel classes. 

\begin{figure}[th]
\centering 
\begin{minipage}{0.23\textwidth}
        \includegraphics[width=\linewidth]{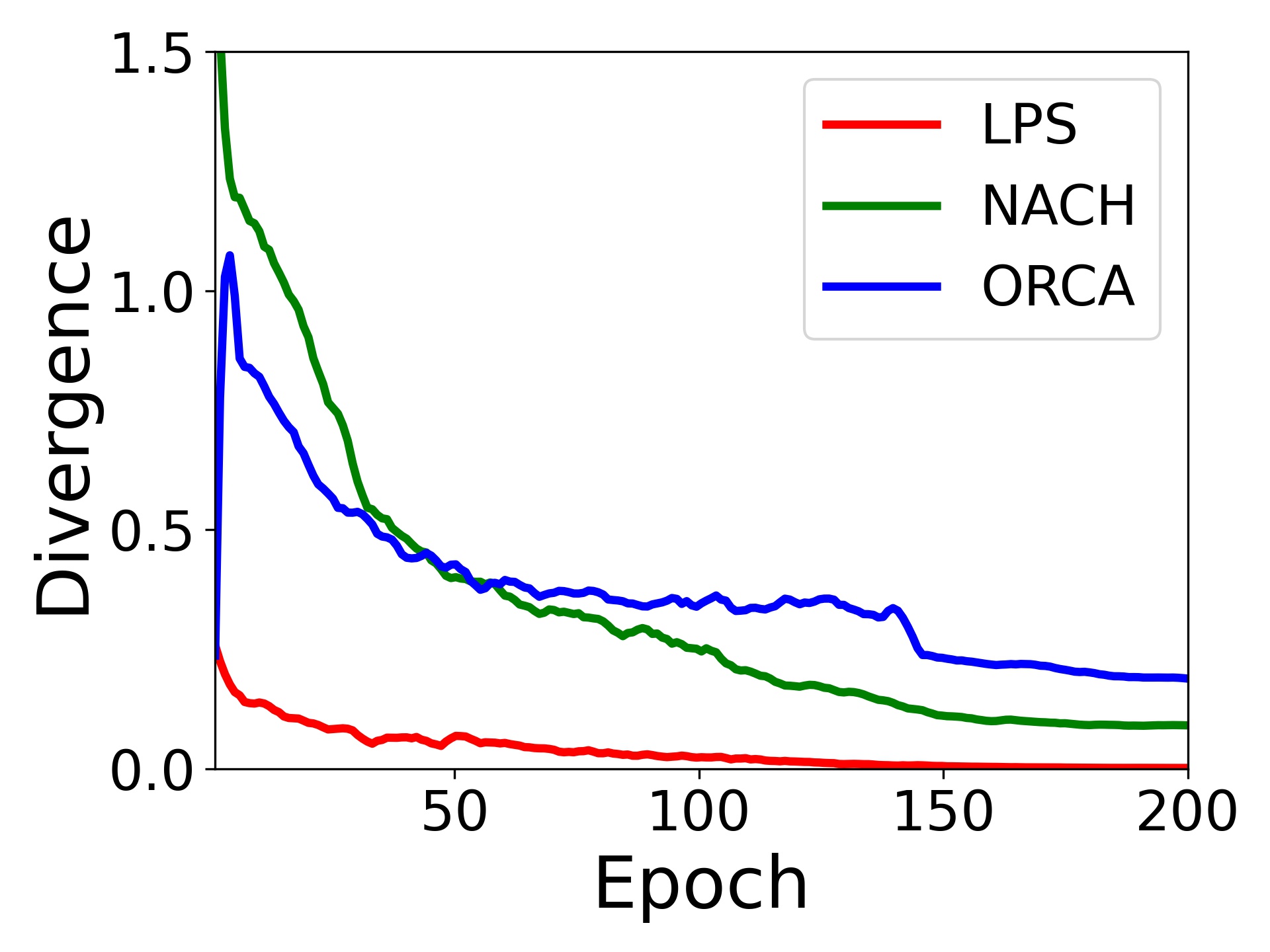}
        \caption*{(a) KL Divergence.}
\end{minipage}
\begin{minipage}{0.23\textwidth}
        \includegraphics[width=\linewidth]{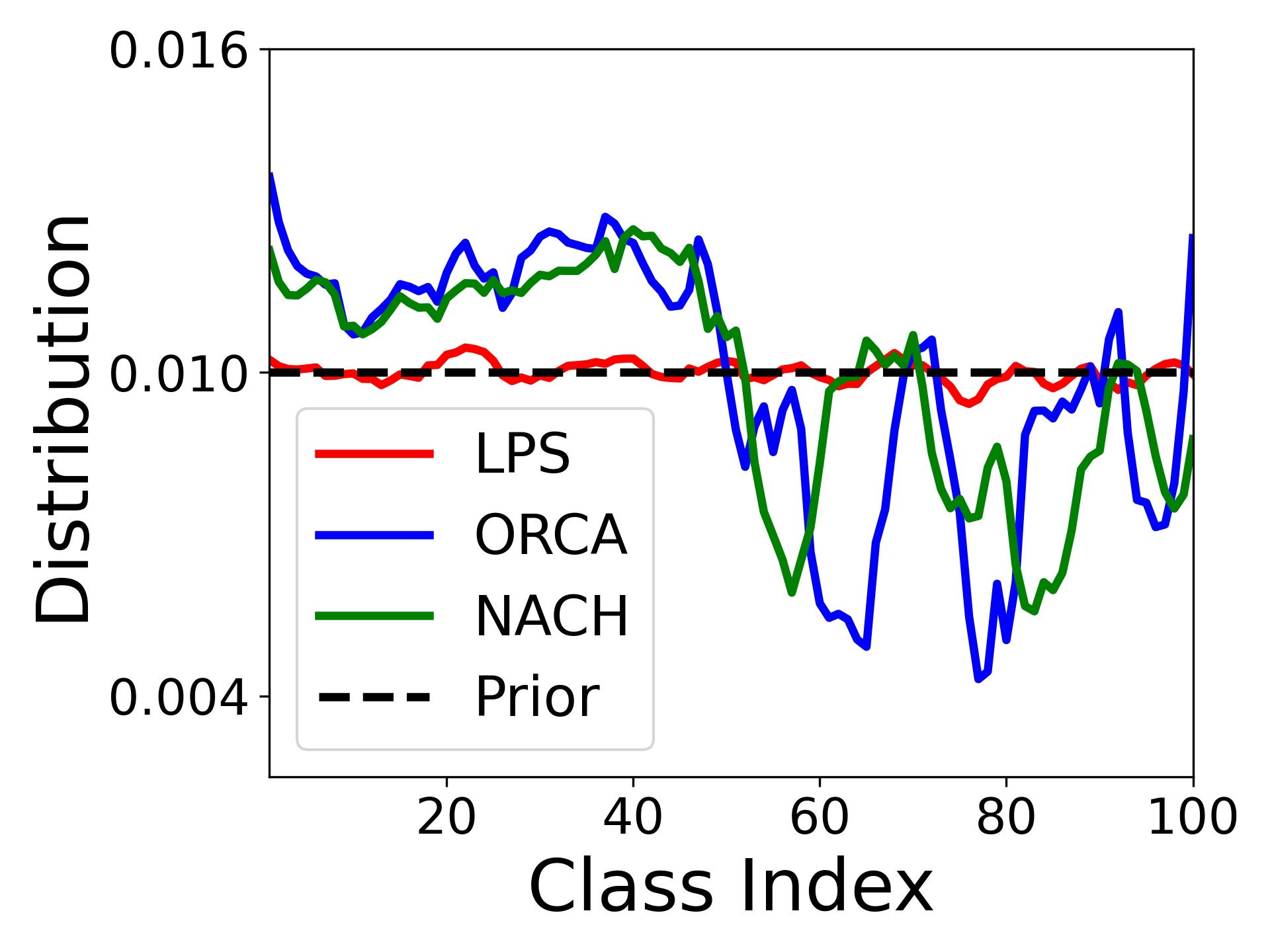}
        \caption*{(b) Class Distribution.}
\end{minipage}
\caption{(a) The KL divergence between the estimated and prior class distributions. (b) The estimated class distribution of the final training iteration. Experiments are conducted on the CIFAR-100 dataset with 50\% seen classes (50\% labeled) and 50\% novel classes.}\label{fig:distcompare}
\end{figure}

\smallskip
\noindent\textbf{Distribution Analysis.}
For further validation of our approach, we present a comprehensive analysis of the KL divergence trend between the estimated and prior class distributions, along with the estimated class distributions at the concluding training iteration for LPS, NACH, and ORCA in Figure~\ref{fig:distcompare}. It can be seen that LPS is able to give more attention to novel classes in the early training stage, in contrast to ORCA and NACH, which exhibit a preference for learning seen classes primarily. This becomes particularly evident in the estimated class distribution at the end of the training, where ORCA and NACH falter in effectively distinguishing novel classes, leading to the erroneous assignment of certain novel class samples. In contrast, the estimated distribution obtained by LPS closely aligns with the prior distribution, consequently fostering effective novel class differentiation.
\begin{figure}[!h]
  \centering
    \subfloat[SimCLR pre-training]{
  \includegraphics[width=0.48\linewidth]{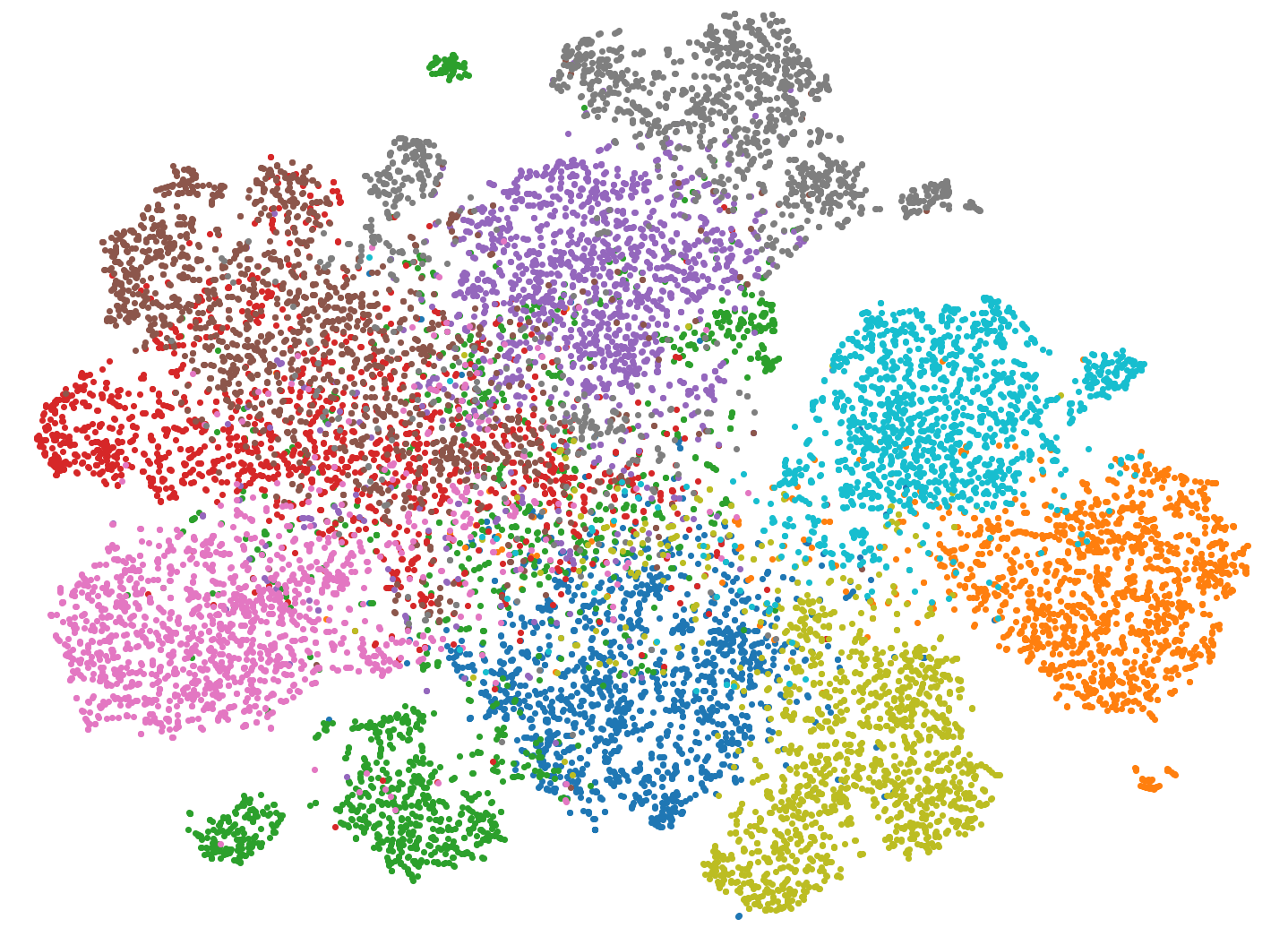}}
  \subfloat[ORCA]{
  \includegraphics[width=0.48\linewidth]{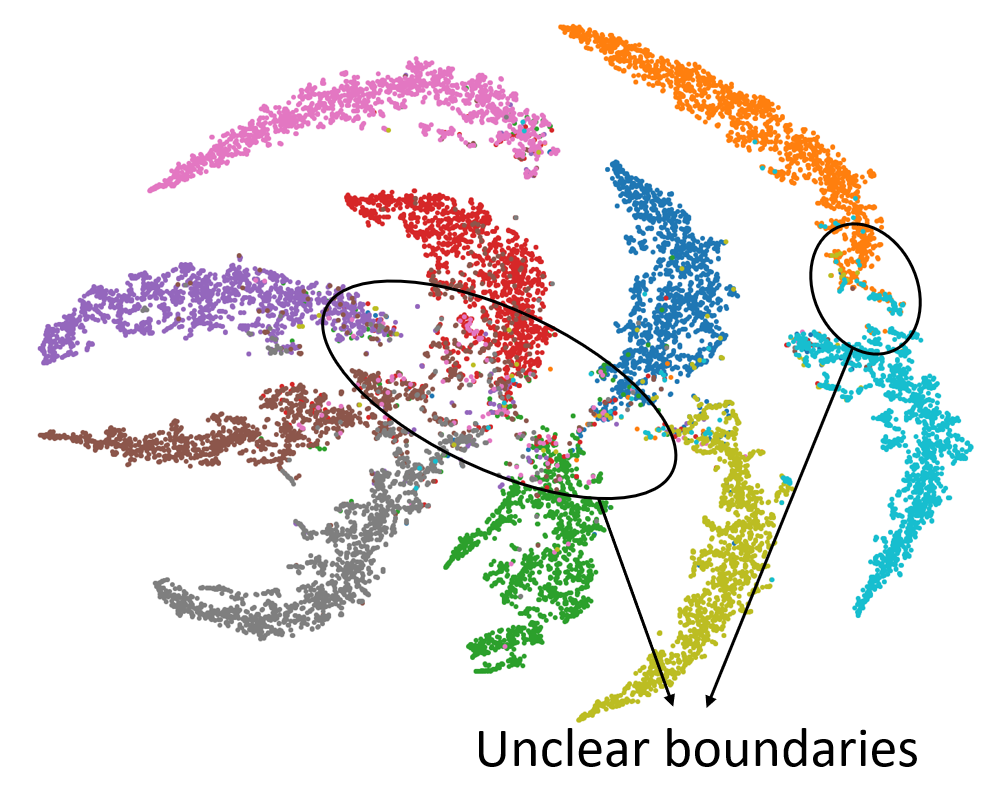}}
    \\
    \subfloat[NACH]{
    \includegraphics[width=0.48\linewidth]{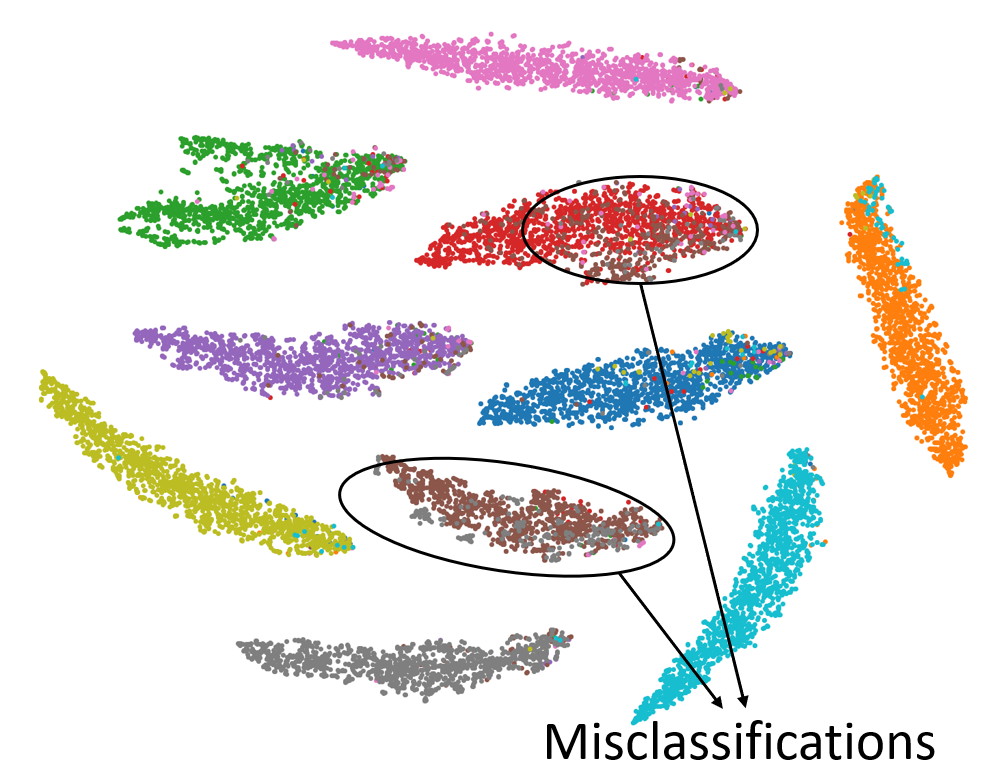}}
    \subfloat[LPS (ours)]{
  \includegraphics[width=0.48\linewidth]{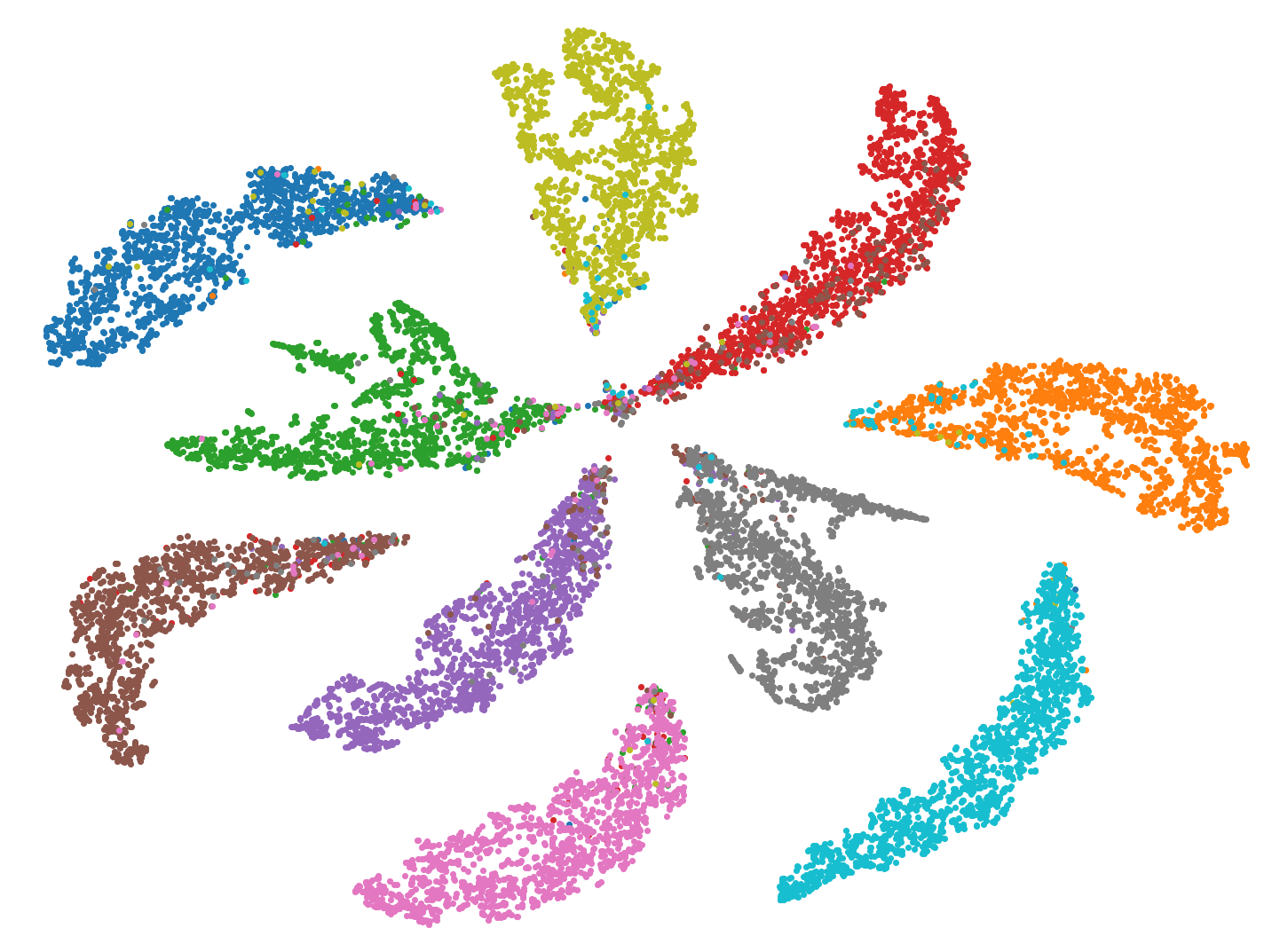}}

  \caption{The t-SNE visualization of features for different methods on CIFAR-10 dataset with 50\% seen classes (50\% labeled) and 50\% novel classes.}
  \label{fig:t_SNE}
\end{figure}

\smallskip
\noindent\textbf{Feature Visualization.}
We employ t-distributed stochastic neighbour embedding (t-SNE) \cite{van2008visualizing} to visualize the learned feature representations through \algo\ and prior methods. The results depicted in Figure~\ref{fig:t_SNE} demonstrate that the representations obtained by LPS yield clearer classification boundaries and exhibit fewer instances of misclassifications. Conversely, ORCA and NACH manifest a higher frequency of misclassifications and unclear boundaries, resulting in comparatively inferior performance in contrast to \algo. In addition, we utilize the metric of normalized mutual information (NMI) to further assess the clustering performance and representations of novel classes. We report the results in Table \ref{tab:NMI} which clearly demonstrate the superiority of LPS. Given that all methods share the same foundation of SimCLR pre-trained backbone, the visualization, and NMI results highlight the efficacy of our approach in enhancing representation learning.

\begin{table}[h]
	\centering
	\begin{tabular}{lccccccc}
		\toprule
          & \multicolumn{3}{c}{\textbf{CIFAR-10}}  &                    
		 & \multicolumn{3}{c}{\textbf{CIFAR-100}} \\
		\textbf{Methods}       & \multicolumn{1}{c}{\textbf{10\%}} & & \multicolumn{1}{c}{\textbf{50\%}} & & \multicolumn{1}{c}{\textbf{10\%}}& & \multicolumn{1}{c}{\textbf{50\%}}\\
		\midrule
            {ORCA} & 73.93 & \quad &80.92 & & 45.26 & \quad &52.10 \\
		{NACH} & 80.07 & \quad &83.44 & &  49.96 & \quad &56.45\\
		{LPS (ours)}  & \textbf{82.41}  & \quad &\textbf{86.45} & & \textbf{51.92} & \quad &\textbf{58.42}\\
		\bottomrule
	\end{tabular}
	\caption{The NMI of novel classes on CIFAR datasets.}
    \label{tab:NMI}
\end{table}

\smallskip
\noindent\textbf{Fine-tuning the Pre-trained Backbone.}
Furthermore, it is noteworthy that all previous OpenSSL methods adopt a practice of freezing the parameters within the first three blocks of the backbone, solely fine-tuning the last block, with the intention of mitigating overfitting. However, such an approach constrains the extent of performance enhancement, as the backbone's parameters remain unmodifiable and unoptimized to better suit downstream tasks. To establish that our method is not susceptible to the overfitting dilemma, we conducted a series of experiments on the CIFAR dataset employing state-of-the-art OpenSSL methods while fine-tuning the backbone. The results are reported in Table \ref{tab:finding1}. 
The experimental results reveal that existing OpenSSL methods manifest modest performance improvement, if any, in comparison to their initial frozen counterparts. In contrast, our proposed method, unaffected by overfitting concerns, consistently yields substantial performance gains across both seen and novel classes. Specifically, the overall accuracy for CIFAR-10 experiences a notable improvement of 2.9\%, while an impressive 6.3\% increase is observed for CIFAR-100. These results underscore the effectiveness of \algo\ in harnessing the additional learnable parameters for further enhancing model performance.


\begin{table}[tp]
	\centering
	\setlength{\tabcolsep}{1.4mm}{
	\begin{tabular}{lccccccc}
		\toprule
          & \multicolumn{3}{c}{\textbf{CIFAR-10}}  &                    
		 & \multicolumn{3}{c}{\textbf{CIFAR-100}} \\
		\textbf{Methods}       & \multicolumn{1}{c}{\textbf{Seen}} & \multicolumn{1}{c}{\textbf{Novel}} & \multicolumn{1}{c}{ \textbf{All}} & & \multicolumn{1}{c}{\textbf{Seen}} & \multicolumn{1}{c}{\textbf{Novel}} & \multicolumn{1}{c}{ \textbf{All}}\\
		\midrule
            {ORCA} & 89.8 &\,  90.8 &  90.5 & \quad &69.4&\, 42.5 &  48.2 \\
            {OpenNCD} & 88.9 &\,  91.8 &  90.8 & \quad &62.2&\, 44.9 &  50.4 \\
		{NACH} & 95.0 &\, 93.3 &  93.9 & \quad &73.8&\,  47.8 &  54.6\\
		{\algo\ (ours)}  & 95.0 &\, 95.5 &  \textbf{95.3} & \quad &72.6&\, 55.5 &  \textbf{60.6} \\
		\bottomrule
	\end{tabular}}
	\caption{Mean accuracy over three runs with removing the limitation of freezing the backbone on CIFAR datasets with 50\% seen classes (50\% labeled) and 50\% novel classes.}
    \label{tab:finding1}
\end{table}

\begin{table}[tp]
	\centering
	\setlength{\tabcolsep}{1.4mm}{
	\begin{tabular}{lccccccc}
		\toprule
          & \multicolumn{3}{c}{\textbf{CIFAR-10}}  &                    
		 & \multicolumn{3}{c}{\textbf{CIFAR-100}} \\
		\textbf{Methods}       & \multicolumn{1}{c}{\textbf{Seen}} & \multicolumn{1}{c}{\textbf{Novel}} & \multicolumn{1}{c}{ \textbf{All}} & & \multicolumn{1}{c}{\textbf{Seen}} & \multicolumn{1}{c}{\textbf{Novel}} & \multicolumn{1}{c}{ \textbf{All}}\\
		\midrule
  	{w/o  $\mathcal{L}_{\text{AM}}$} & 90.6 &\, 89.4 &  89.8 & \quad &67.1&\, 46.2 &  52.7\\

            {w/o  $\mathcal{L}_{\text{PC}}$} & 89.7 &\, 89.1 &  89.3 & \quad &60.4&\, 48.1 &  51.8\\
            		{w/o  $\mathcal{L}_{\text{UC}}$} & 91.3 &\, 88.2 &  89.2 & \quad &64.7&\,  42.3 &  48.5\\
   {w/o  $\mathcal{R}_{\text{Entropy}}$} & 90.8 &\,  55.4 &  66.2 & \quad &73.4&\, 28.8 &  30.8 \\
		{\algo\ (ours)}  & 90.2 &\, 93.4 &  \textbf{92.4} & \quad &64.5&\, 49.9 &  \textbf{54.3} \\
		\bottomrule
	\end{tabular}}
	\caption{Accuracy when removing key components of our method. We report the average accuracy over three runs on CIFAR datasets with 50\% seen classes (50\% labeled) and 50\% novel classes.}
    \label{tab:ablation1}
\end{table}

\smallskip
\noindent\textbf{Ablation Analysis.}
Moreover, we conduct a comprehensive analysis of the contributions of distinct components in our approach. The objective function of \algo\ comprises the adaptive margin loss ($\mathcal{L}_{\text{AM}}$), the pseudo-label contrastive clustering loss ($\mathcal{L}_{\text{PC}}$), the unsupervised contrastive learning loss ($\mathcal{L}_{\text{UC}}$), and the entropy regularizer ($\mathcal{R}_{\text{Entropy}}$). Concretely, the ablation study is mainly conducted by removing each term individually from the objective function except for the adaptive margin which is replaced by a standard cross-entropy. As observed in Table \ref{tab:ablation1}, the removal of any components leads to performance degradation. The substantial drop in novel performance after removing the entropy regularizer highlights its significant role in the process of novel class discovery. Moreover, the utilization of both pseudo-label contrastive loss and adaptive margin loss substantially improves the accuracy of novel classes.

\section{Conclusion}
In this study, we present Learning Pace Synchronization (\algo), a potent solution tailored to the OpenSSL problem. \algo\ introduces an adaptive margin loss to effectively narrow the learning pace gap that exists between seen and novel classes. Moreover, we formulate a pseudo-label contrastive clustering loss to augment the process of novel class discovery. Extensive evaluation is conducted across three benchmark datasets with distinct quantities of labeled data. We also discover that the conventional practice of freezing the self-supervised pre-trained backbone hinders the generalization performance. We hope our work can inspire more efforts towards this realistic setting.
\section{Acknowledgments}
This work was supported by the National Science Foundation of China (62206049, 62225602), and the Big Data Computing Center of Southeast University.

\section{Supplementary Material for LPS}
\subsection{Parameter Sensitivity Analysis}
Parameters $\eta_1$ and $\eta_2$ define the weight of the pseudo-label contrastive clustering loss and the unsupervised contrastive learning loss, respectively. We conduct several experiments on the CIFAR datasets with various values of $\eta_1$ and $\eta_2$ to assess the performance of our method LPS, and the results are shown in Table \ref{tab:alph}. By further adjusting the parameters $\eta_1$ and $\eta_2$, our method LPS displays great robustness and promising results. We also provide detailed analysis for $\lambda_{novel}$. In particular, we set $\lambda_{novel}=0.4+ \{0.3,0.35,0.4,0.45,0.5\} \times \frac{t}{T}$ and the results are reported in Table \ref{tab:thres}. We find that higher values of $\lambda_{novel}$ achieve better performance on seen classes. Intuitively,  higher values of $\lambda_{novel}$ will pseudo-label less novel unlabeled samples and further let $\mathcal{L}_{\text{AM}}$ give more importance to seen classes.

\begin{table}[!h]
	\centering
	\setlength{\tabcolsep}{1.5mm}{
	\begin{tabular}{cccccccc}
		\toprule
          & \multicolumn{3}{c}{\textbf{CIFAR-100}}  &                    
		 & \multicolumn{3}{c}{\textbf{CIFAR-100}} \\
		\textbf{$\eta_1$}       & \multicolumn{1}{c}{\textbf{Seen}} & \multicolumn{1}{c}{\textbf{Novel}} & \multicolumn{1}{c}{ \textbf{All}} &  \textbf{$\eta_2$} & \multicolumn{1}{c}{\textbf{Seen}} & \multicolumn{1}{c}{\textbf{Novel}} & \multicolumn{1}{c}{ \textbf{All}}\\
		\midrule
            {0.6} & 63.5 &\,  48.4 &  53.1 &  {0.6} &64.7&\, 50.8 &  \textbf{55.1} \\
		{0.8} & 63.7 &\, 49.3 &  53.7 & {0.8} &64.4&\,  49.8 &  53.6\\
            {1.0} & 64.5 &\, 49.9 &  54.3 & {1.0} &64.5&\, 49.9 &  54.3\\
		{1.2} & 64.2 &\, 50.2&  \textbf{54.4} & {1.2} &63.7&\, 49.8 &  54.0\\
		{1.4}  & 64.1 &\, 48.8 &  53.5 & {1.4} &63.2&\, 50.6 &  54.4 \\
		\bottomrule
	\end{tabular}}
 \vspace{-0.5em}
	\caption{Accuracy with various values of $\eta_1$ and $\eta_2$.}
    \label{tab:alph}
    \vspace{-1em}
\end{table}

\begin{table}[!h]
	\centering
	\setlength{\tabcolsep}{1.5mm}{
	\begin{tabular}{cccccccc}
		\toprule
          & \multicolumn{3}{c}{\textbf{CIFAR-10}}  &                    
		 & \multicolumn{3}{c}{\textbf{CIFAR-100}} \\
		\textbf{${\lambda}_{novel}$}       & \multicolumn{1}{c}{\textbf{Seen}} & \multicolumn{1}{c}{\textbf{Novel}} & \multicolumn{1}{c}{ \textbf{All}} & & \multicolumn{1}{c}{\textbf{Seen}} & \multicolumn{1}{c}{\textbf{Novel}} & \multicolumn{1}{c}{ \textbf{All}}\\
		\midrule
            {0.70} & 89.9 &\,  92.2 &  91.5 & \quad &62.9&\, 49.3 &  53.4 \\
		{0.75} & 89.8 &\, 93.5 &  92.3 & \quad &63.9&\,  49.0 &  53.5\\
            {0.80} & 90.2 &\, 93.4 &  \textbf{92.4} & \quad &64.5&\, 49.9 &  54.3\\
		{0.85} & 90.5 &\, 91.7 &  91.3 & \quad &64.8&\, 48.0 &  53.2\\
		{0.90}  & 90.9 &\, 90.3 &  90.5 & \quad &65.2&\, 49.6 &  \textbf{54.4} \\
		\bottomrule
	\end{tabular}}
 \vspace{-0.5em}
	\caption{Accuracy with various values of ${\lambda}_{novel}$.}
    \label{tab:thres}
\end{table}

Recall that $C$ defined in $\mathcal{L}_{\text {AM}}$ is a tunable parameter to control the maximum margin. Table \ref{tab:SensitivityAnalysis} displays the results of LPS under varying C conditions. Intuitively, increasing the $C$ will lead to a faster alignment between the predicted distribution and the prior distribution. From the results of $C=20$ in the Table \ref{tab:SensitivityAnalysis}, if the alignment is too fast, the model may balance the learning pace between seen and novel classes by assigning incorrect pseudo-labels. Meanwhile, if the alignment is too slow, the margin mechanism does not effectively bias the model towards novel classes, which is reflected in the results of $C=1$ and $C=5$ in the Table \ref{tab:SensitivityAnalysis}. 

\begin{table}[!h]
	\centering
	\setlength{\tabcolsep}{1.5mm}{
	\begin{tabular}{cccccccc}
		\toprule
          & \multicolumn{3}{c}{\textbf{CIFAR-10}}  &                    
		 & \multicolumn{3}{c}{\textbf{CIFAR-100}} \\
		\textbf{$C$}       & \multicolumn{1}{c}{\textbf{Seen}} & \multicolumn{1}{c}{\textbf{Novel}} & \multicolumn{1}{c}{ \textbf{All}} & & \multicolumn{1}{c}{\textbf{Seen}} & \multicolumn{1}{c}{\textbf{Novel}} & \multicolumn{1}{c}{ \textbf{All}}\\
		\midrule
            {1} & 90.5 &\,  90.6 &  90.5 & \quad &66.6&\, 45.3 &  51.9 \\
		{5} & 90.3 &\, 91.9 &  91.4 & \quad &64.6&\,  47.4 &  52.7\\
            {10} & 90.2 &\, 93.4 &  \textbf{92.4} & \quad &64.5&\, 49.9 &  54.3\\
		{15} & 90.2 &\, 93.3 &  92.2 & \quad &63.2&\, 50.8 &  \textbf{54.6}\\
		{20}  & 90.2 &\, 91.9 &  91.3 & \quad &62.4&\, 48.7 &  52.9 \\
		\bottomrule
	\end{tabular}}
	\caption{Accuracy with various values of $C$.}
    \label{tab:SensitivityAnalysis}
\end{table}

The temperature parameter $\tau$ within $\mathcal{L}_{\text{PC}}$ and $\mathcal{L}_{\text{UC}}$ is used to adjust the measurement of similarity between samples. Increasing the temperature will lead to a flatter representation space for samples. Conversely, decreasing it results in a sharper space. We conduct a series of experiments by setting various temperature values. Table \ref{tab:SensitivityAnalysis2} depicts that alterations in $\tau$ do not exert a pronounced impact on performance and $\tau=0.4$ yields the best performance. 


\begin{table}[!h]
	\centering
	\setlength{\tabcolsep}{1.5mm}{
	\begin{tabular}{cccccccc}
		\toprule
          & \multicolumn{3}{c}{\textbf{CIFAR-10}}  &                    
		 & \multicolumn{3}{c}{\textbf{CIFAR-100}} \\
		\textbf{$\tau$}       & \multicolumn{1}{c}{\textbf{Seen}} & \multicolumn{1}{c}{\textbf{Novel}} & \multicolumn{1}{c}{ \textbf{All}} & & \multicolumn{1}{c}{\textbf{Seen}} & \multicolumn{1}{c}{\textbf{Novel}} & \multicolumn{1}{c}{ \textbf{All}}\\
		\midrule
            {0.2} & 85.3 &\,  92.7 &  90.3 & \quad &61.4&\, 48.8 &  52.6 \\
		{0.3} & 89.2 &\, 92.2 &  91.2 & \quad &63.2&\,  49.2 &  53.4\\
            {0.4} & 90.2 &\, 93.4 &  \textbf{92.4} & \quad &64.5&\, 49.9 &  \textbf{54.3}\\
		{0.5} & 90.9 &\, 92.3 &  91.8 & \quad &64.4&\, 47.2 &  52.5\\
		{0.6}  & 91.0 &\, 91.3 &  91.2 & \quad &63.8&\, 46.5 &  51.8 \\
		\bottomrule
	\end{tabular}}
	\caption{Accuracy with various values of $\tau$.}
    \label{tab:SensitivityAnalysis2}
\end{table}



\subsection{Additional Results}
\begin{figure}[!h]
\centering 
\begin{minipage}{0.23\textwidth}
        \includegraphics[width=\linewidth]{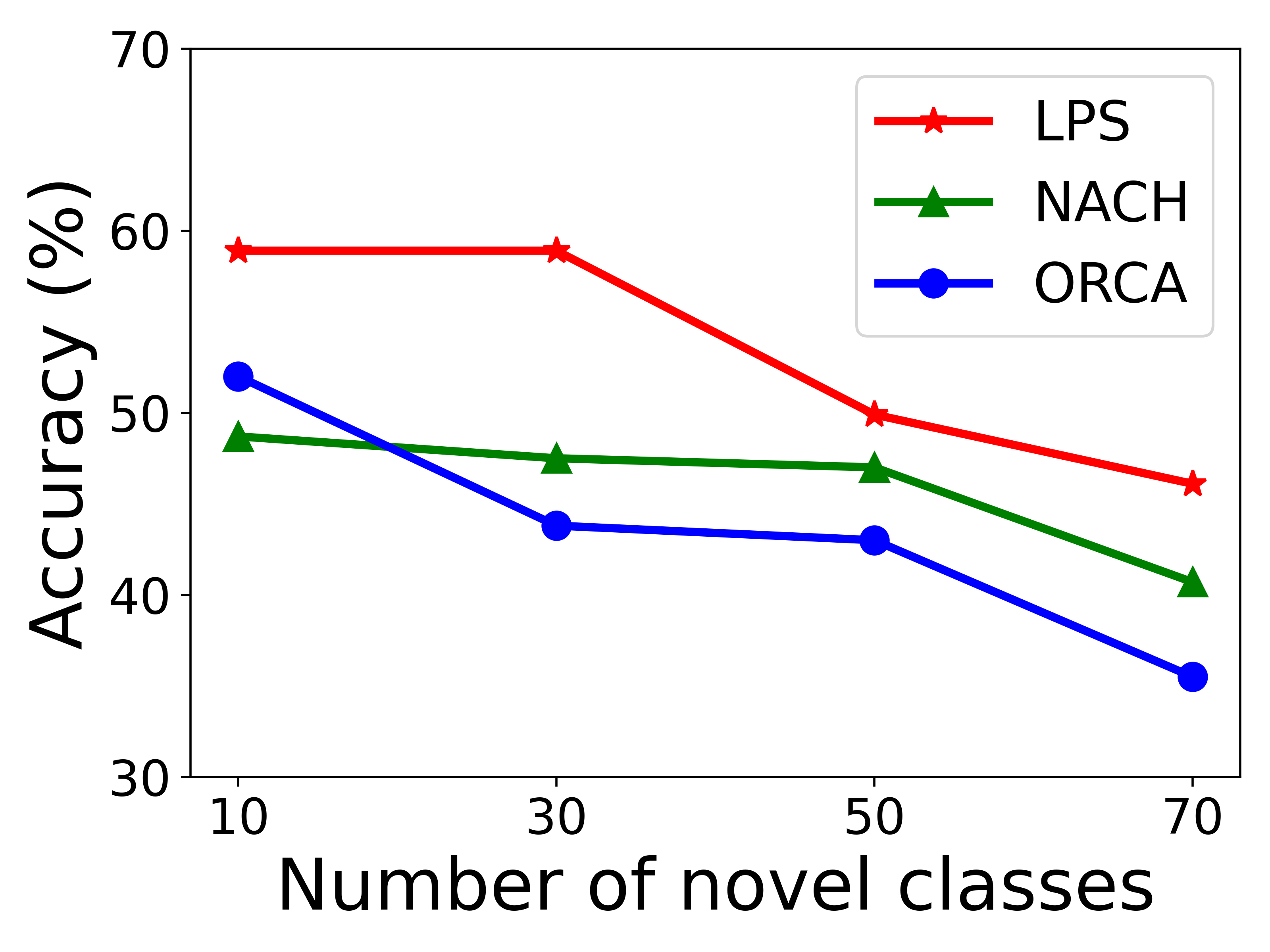}
        \caption*{(a) Novel Class Accuracy.}
\end{minipage}
\begin{minipage}{0.238\textwidth}
        \includegraphics[width=\linewidth]{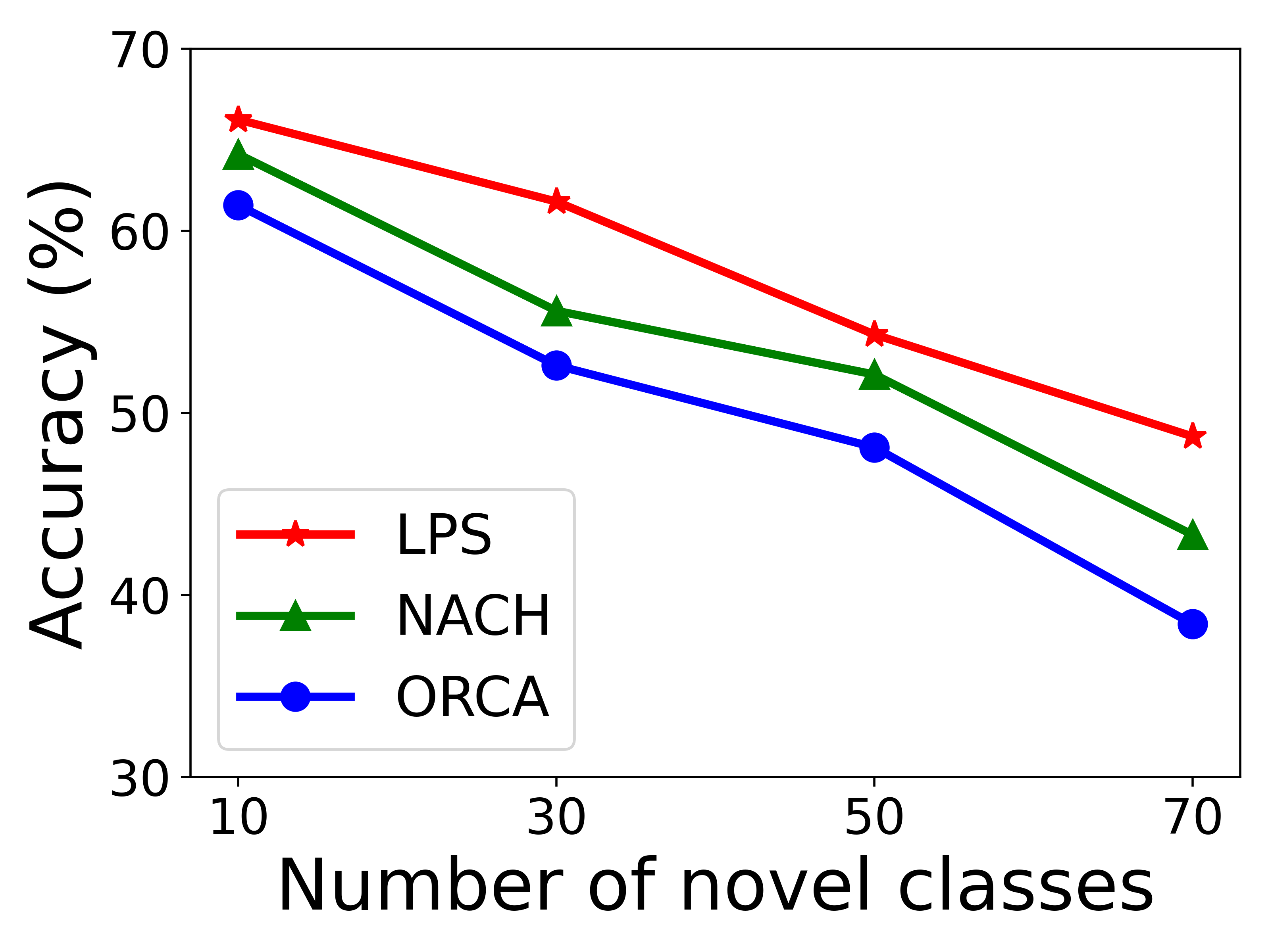}
        \caption*{(b) Overall Accuracy.}
\end{minipage}
\begin{minipage}{0.23\textwidth}
        \includegraphics[width=\linewidth]{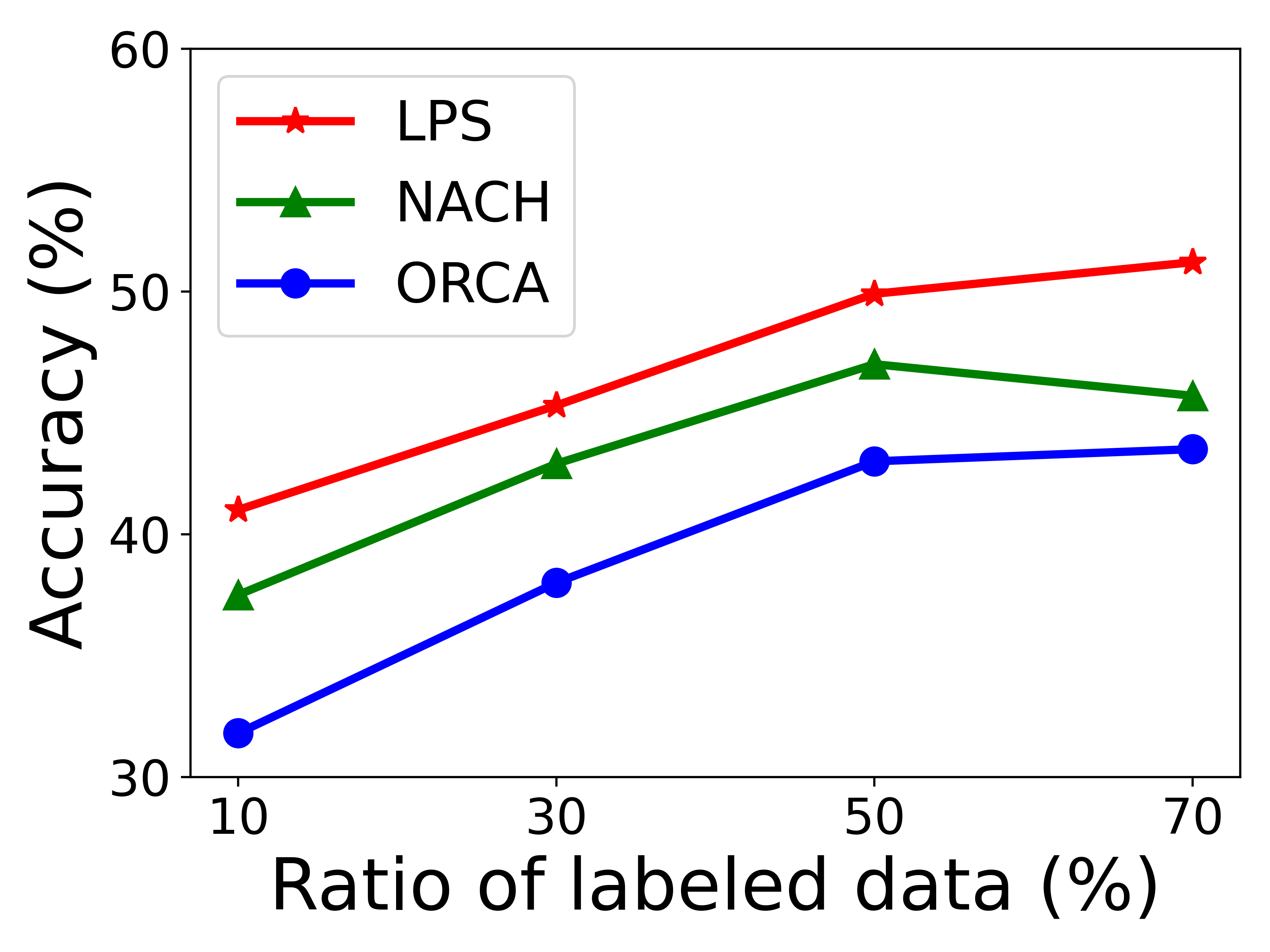}
        \caption*{(c) Novel Class Accuracy.}
\end{minipage}
\begin{minipage}{0.238\textwidth}
        \includegraphics[width=\linewidth]{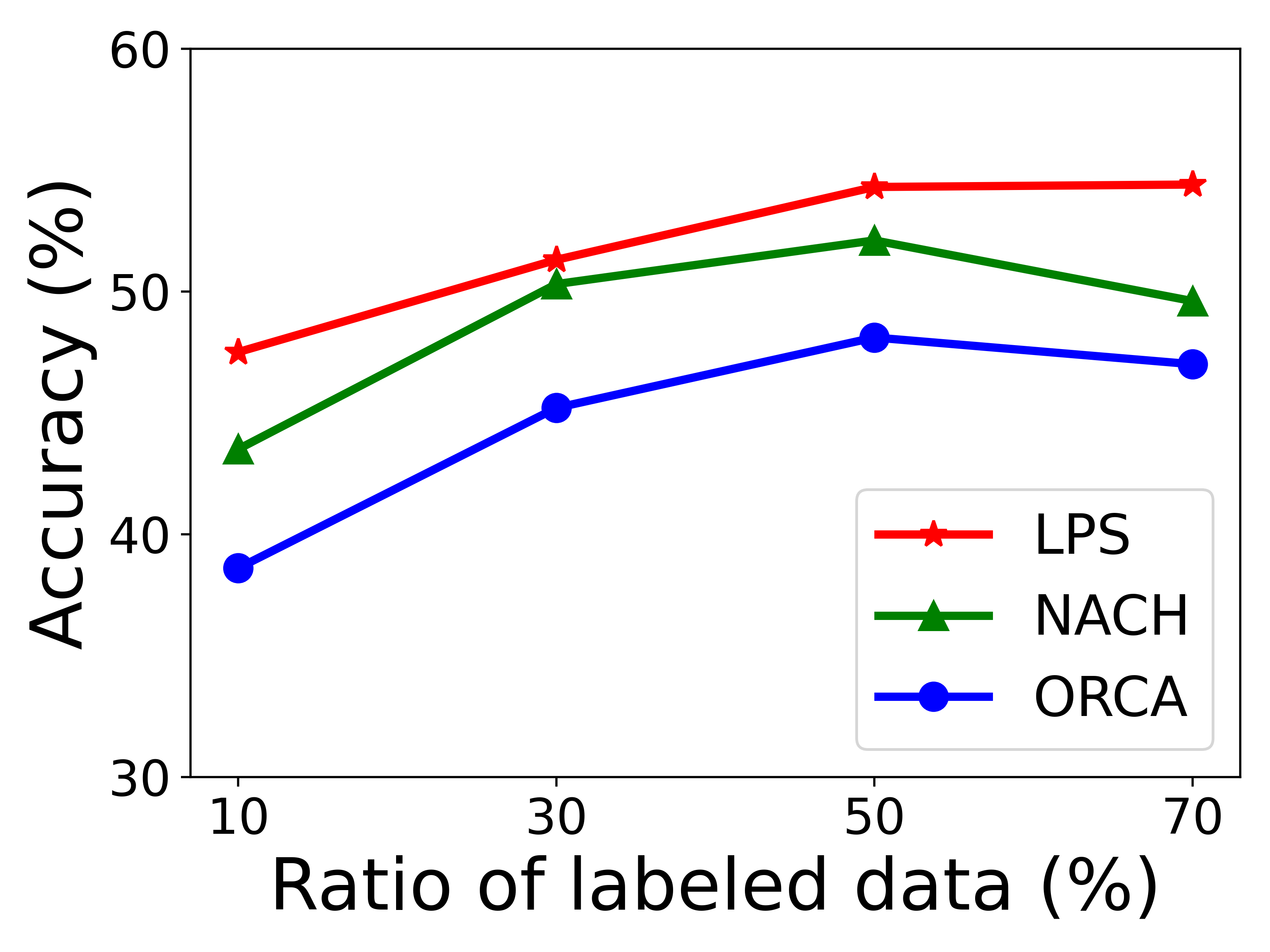}
        \caption*{(b) Overall Accuracy.}
\end{minipage}
\caption{(a) The novel class accuracy and (b) overall accuracy with different numbers of novel classes. (c) The novel class accuracy and (d) overall accuracy with different ratios of labeled data.}
\label{fig:changes}
\end{figure}

In addition, we conduct more experiments to validate the robustness of the proposed method. We first conduct a series experiments on CIFAR-100 dataset with \textbf{different numbers of novel classes}, and the results are reported in the Figure~\ref{fig:changes} (a) and (b). Further, we conduct a series experiments on CIFAR-100 dataset with \textbf{different ratios of labeled data}, and the results are shown in the Figure~\ref{fig:changes} (c) and (d). From the results, it can be clearly seen that LPS consistently outperforms ORCA and NACH across all settings. In Figure~\ref{fig:changes} (c) and (d), both ORCA and NACH deteriorate their performance when the ratio of labeled data reaches 70\%, while LPS is able to yield further improvement.


To further evaluate the performance when \textbf{fine-tuning the pre-trained backbone}, we conduct a series of experiments on the CIFAR dataset with 50\% seen classes (10\% labeled) and 50\% novel classes. From Table \ref{tab:nofreeeze}, we can see that both ORCA and NACH show significant declines (over 10\% overall accuracy), while our method LPS maintains high performance on CIFAR-100 and shows further improvements on CIFAR-10, which further verifies that LPS is not susceptible to the overfitting dilemma.
\begin{table}[!h]
	\centering
	\setlength{\tabcolsep}{1.4mm}{
	\begin{tabular}{lccccccc}
		\toprule
          & \multicolumn{3}{c}{\textbf{CIFAR-10}}  &                    
		 & \multicolumn{3}{c}{\textbf{CIFAR-100}} \\
		\textbf{Methods}       & \multicolumn{1}{c}{\textbf{Seen}} & \multicolumn{1}{c}{\textbf{Novel}} & \multicolumn{1}{c}{ \textbf{All}} & & \multicolumn{1}{c}{\textbf{Seen}} & \multicolumn{1}{c}{\textbf{Novel}} & \multicolumn{1}{c}{ \textbf{All}}\\
		\midrule
            {ORCA} & 78.1 &\,  80.7 &  79.5 & \quad &39.7&\,23.7 &  29.6 \\
		{NACH} & 87.0 &\, 82.5 &  80.8 & \quad &51.0&\,  28.9 &  37.1\\
		{LPS(ours)}  & 88.6 &\, 91.6 &  \textbf{90.2} & \quad &57.0&\, 39.3 &  \textbf{47.3} \\
		\bottomrule
	\end{tabular}}
	\caption{Accuracy without freezing the backbone on CIFAR datasets.}
    \label{tab:nofreeeze}
\end{table}
\end{document}